\documentclass[10pt,twocolumn,letterpaper]{article}

\usepackage{cvpr}              
\usepackage{multirow}
\usepackage{adjustbox}
\usepackage[accsupp]{axessibility}
\usepackage[dvipsnames]{xcolor}

\definecolor{cvprblue}{rgb}{0.21,0.49,0.74}
\usepackage[pagebackref,breaklinks,colorlinks,citecolor=cvprblue]{hyperref}
\usepackage{colortbl}

\def\paperID{8639} 
\def\confName{CVPR}
\def\confYear{2024}

\title{Solving Masked Jigsaw Puzzles with Diffusion Vision Transformers
}

\author{Jinyang Liu\\
Northeastern University\\
Boston MA\\
{\tt\small liu.jinyan@northeastern.edu}
\and
Wondmgezahu Teshome\\
Northeastern University\\
Boston MA\\
{\tt\small teshome.w@northeastern.edu}
\and
Sandesh Ghimire\\
Qualcomm \\
San Diego CA\\
{\tt\small drsandeshghimire@gmail.com}
\and
Mario Sznaier\\
Northeastern University\\
Boston MA\\
{\tt\small msznaier@coe.northeastern.edu}
\and
Octavia Camps\\
Northeastern University\\
Boston MA\\
{\tt\small camps@coe.northeastern.edu}
}

\begin{document}
\maketitle
\begin{abstract}


Solving image and video jigsaw puzzles poses the challenging task of rearranging image fragments or  video frames from unordered sequences to restore meaningful images and video sequences.
Existing approaches often hinge on discriminative models tasked with predicting either the absolute positions of puzzle elements or the permutation actions applied to the original data. Unfortunately, these methods face limitations in effectively solving puzzles with a large number of elements.
In this paper, we propose JPDVT, an innovative approach that harnesses diffusion transformers to address this challenge. Specifically, we generate positional information for image patches or video frames, conditioned on their underlying visual content. This information is then employed to accurately assemble the puzzle pieces in their correct positions, even in scenarios involving missing pieces.
Our method achieves state-of-the-art performance on several datasets.

\end{abstract}    
\section{Introduction}
\label{sec:intro}

 \textit{Image and video jigsaw puzzle solving} involves the challenging task of reassembling image fragments or reshuffling video frames from unordered sequences into meaningful images and video sequences. This problem holds significant importance across various domains, ranging from image editing, biology, archaeology, document or photograph restoration, to photo sequencing \cite{bridger2020solving}. 



Past methods for solving jigsaw puzzles have commonly relied on discriminative networks, particularly in classifying the positions of image patches or video frames \cite{noroozi2016unsupervised, kim2019self}. Typically, these approaches involve feature extraction using a Convolutional Neural Network (CNN) backbone, followed by pairwise feature combinations before classification. The classifier then operates either by classifying permutations applied to the raw data or by directly predicting the positions of unordered elements. The primary emphasis in these methods is on training the feature  backbone network to extract highly distinguishable positional features during the jigsaw puzzle-solving process.

\begin{figure}[tb]
\centering
\includegraphics[width=0.48\textwidth]{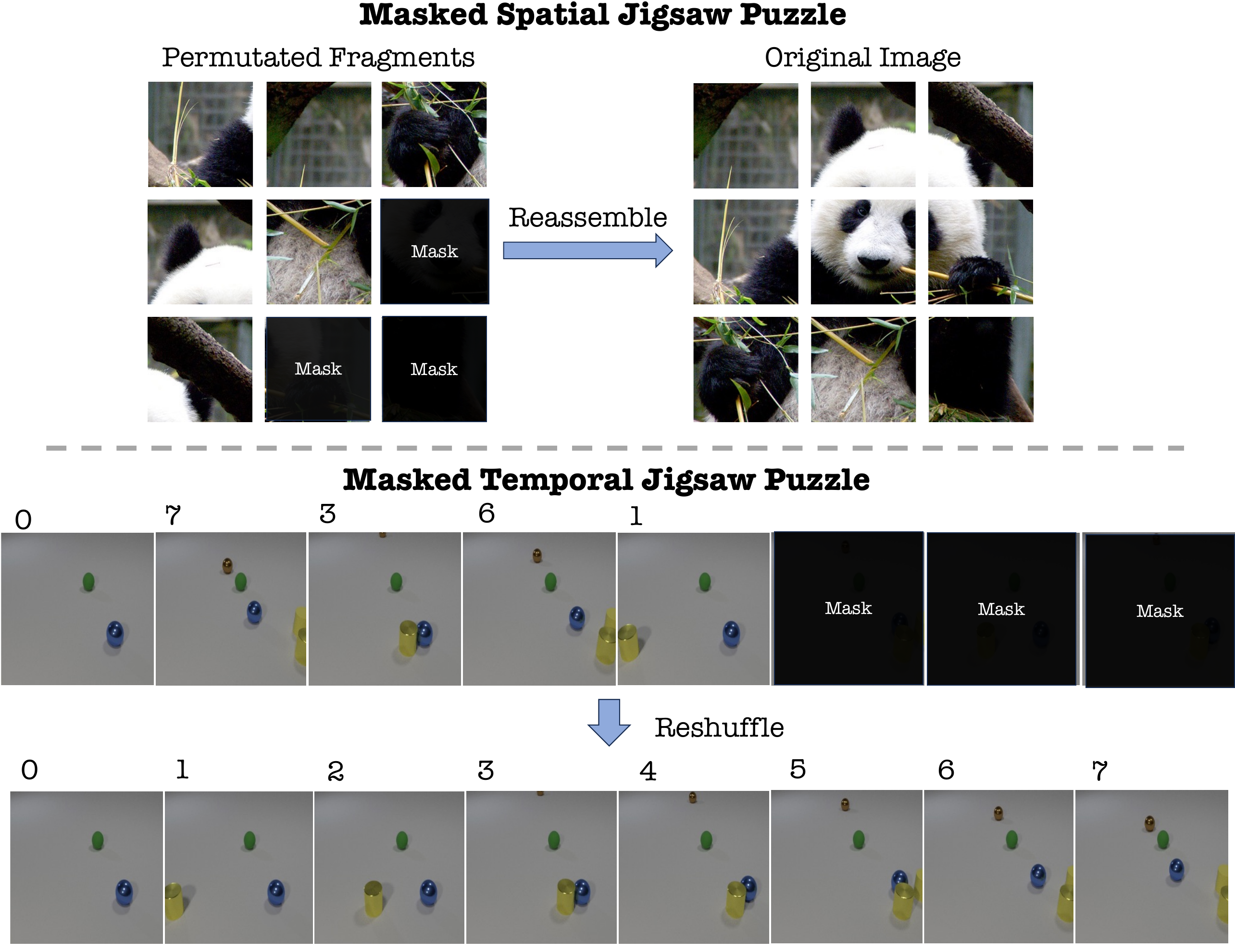}
\caption{Top: given unordered image fragments, some of them masked, we want to reconstruct the original image. Bottom: given shuffled video frames, some of them  masked, we want to reconstruct the original video. }\label{fig_concept}
\vspace{-0.5cm}
\end{figure}

\begin{figure}[tb]
\centering
\includegraphics[width=0.45\textwidth]{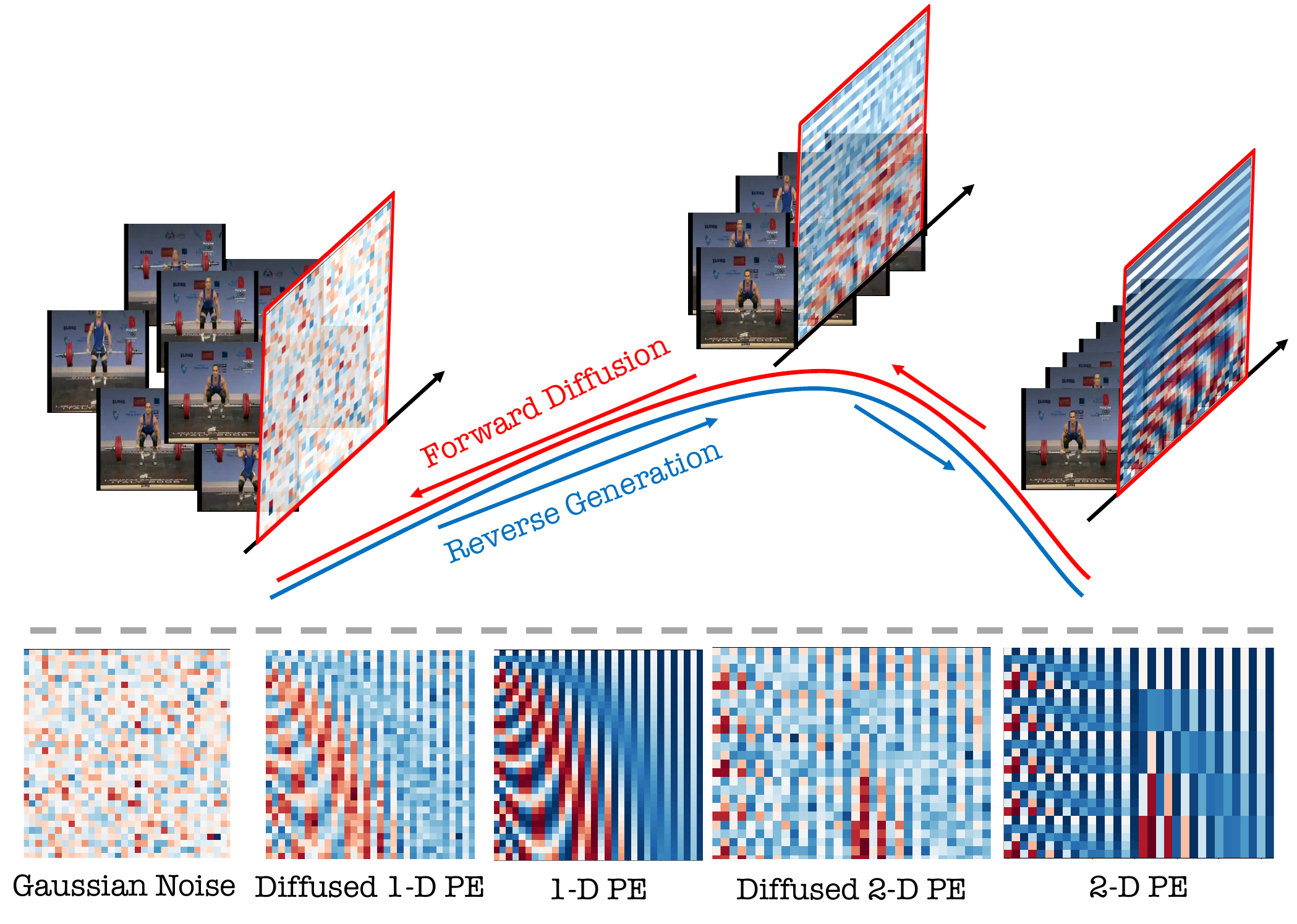}
\caption{Top. Each piece has a positional encoding and an embedding of its  visual content. The forward diffusion step (highlighted in the red box), gradually adds noise  to the positional encodings. The reverse generation reconstructs the  positional encodings, conditioned by the provided visual content. Bottom.  Samples of 1D and 2D positional encodings, for video frames and image tiles, both in their original form and with added noise. }\label{fig_pe}
\vspace{-0.5cm}
\end{figure}

\begin{figure*}[tb]
\centering
\includegraphics[width=1\textwidth]{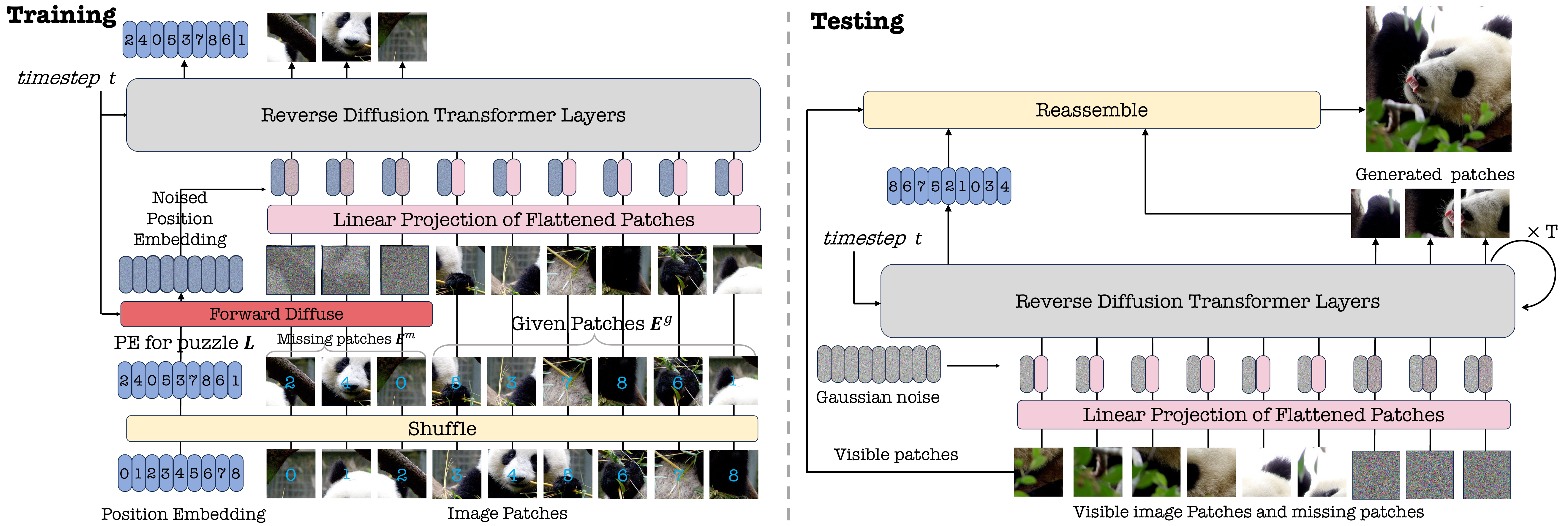}
\caption{Left: Architecture employed to train the proposed diffusion transformer model. Image patches, accompanied by their corresponding position encoding tokens, undergo permutation in a consistent manner. During forward diffusion, noise is introduced to all positional encoding tokens and a subset of embedded patch tokens. The transformer is trained to execute a reverse diffusion process, mitigating the introduced noise.
Right: Inference process. During inference, the observed patches are padded with Gaussian noise to fill the entire image. Linearly embedded patch tokens are then concatenated with positional encoding tokens initialized with Gaussian noise. The transformer model is designed to reconstruct the visual content of the missing patches and to determine the positions for all patches, simultaneously.}\label{fig_arch}
\end{figure*}

However, these approaches face challenges when dealing with a large number of elements and when handling a variable number of elements. Furthermore, they often prioritize learning order at the expense of capturing the entire structure of the data, making them less suitable for solving jigsaw puzzles when data elements are missing.

This paper introduces  {\bf J}igsaw {\bf P}uzzles with  {\bf D}iffusion {\bf V}ision {\bf T}ransformers (JPDVT),  a novel approach to tackle these challenges through the utilization of  a conditional diffusion denoising probabilistic model (CDDPM). The CDDPM is designed to learn the positions of all the elements and generate the missing pieces, guided by the visual context of the available pieces, as illustrated in Figs.~\ref{fig_concept} and \ref{fig_pe}.  Rigorous testing on diverse image and video datasets validates its superior performance compared to state-of-the-art models, underscoring the effectiveness of our proposed approach. The main contributions of this work are:



\begin{enumerate}
\item A simple unified design to solve both image and temporal puzzles with large number of pieces.  The proposed model can handle missing data, finding the correct positions of the available pieces and generating plausible  reconstructs for the missing ones. 
\item State-of-the-art performance on several image and video datasets. Our code and pre-trained models are available at our GitHub project page.\footnote{\url{https://github.com/JinyangMarkLiu/JPDVT}}
\end{enumerate}

\section{Related Work}

\label{sec:relatedworks}

Solving image and video jigsaw puzzles 
have the common objective of learning (spatial or temporal) positional information from unordered elements.


\textbf{Spatial Jigsaw Puzzles:}
Most existing approaches revolve around the reassembly of square puzzle pieces, with a focus on recreating the original image using non-overlapping components. Existing state-of-the-art jigsaw puzzle solvers can be broadly categorized into two streams.

The first stream, represented by algorithms like \cite{noroozi2016unsupervised,paumard2018image,wei2019iterative,paumard2020deepzzle,song2023siamese}, leverages feature extraction or image understanding networks for encoding visual content. Subsequently, these methods engage in fragment reassembly, demonstrating effectiveness in handling fragments with large eroded gaps.

The second stream, as exemplified by \cite{li2021jigsawgan,bridger2020solving}, initiates the process with fragment generation and subsequent reassembly. A neural network serves as a discriminator to verify the correct placement of fragments, yet these methods are generally limited to addressing small eroded gaps.

While  most of the prior works use CNNs, owing to their downsampling capabilities that facilitate classification, our paper advocates for using  transformer models due to their equivariance to shuffling transformations.

In terms of reassembly strategies, traditional approaches include brute force \cite{paumard2018image} or greedy methods \cite{bridger2020solving}. However, these methods face scalability challenges with large-scale puzzles due to their exponentially growing complexity. Some studies propose using classifiers to predict permutations \cite{noroozi2016unsupervised} or fragment positions \cite{wei2019iterative}. Other sophisticated approaches involve Dijkstra’s algorithm with graph-cut \cite{paumard2020deepzzle}, while a recent work \cite{song2023siamese} employs reinforcement learning as the reassembly strategy. Additionally, \cite{song2023solving} utilized a genetic algorithm to tackle the problem. In contrast, our paper introduces the novel use of diffusion generative models for reassembly, leveraging positional information learned from visual patterns encoded by a transformer model. Moreover, our approach extends beyond conventional practices, demonstrating the capability to solve jigsaw puzzles even in the presence of missing elements.

\textbf{Temporal Puzzles:}
 Several recent works \cite{wang2022video,lee2017unsupervised, kim2019self,xu2019self,misra2016shuffle} propose using sorting temporal sequences as a pretext  tasks for self-supervised learning for applications such as action recognition and video anomaly detection. They  employ CNNs as video encoders to predict the order of shuffled videos directly. However, these methods, while effective for specific applications, are not explicitly designed to address the broader challenges of solving temporal puzzles, particularly on larger temporal scales.

In contrast, our work draws inspiration from the paradigm set by \cite{basha2012photo, moses2013space, dekel2014photo, dicle2016solving}, where the emphasis lies on explicitly reshuffling temporal sequences akin to a photo sequencing problem. Unlike method \cite{dicle2016solving}, which learned permutation matrices by solving optimization problems enforcing low-rank structures on shuffled temporal sequences, our approach transcends the limitations of optimization-based techniques. These methods face constraints on sequence length and lose the ability to capture spatial coherence due to significant dimensionality reduction.

This work moves beyond solving purely temporal jigsaw puzzles and tackles a more intricate problem—reshuffling temporal sequences with temporal masks. Specifically, it aims to reconstruct a shuffled temporal sequence or video with missing frames, posing a novel and challenging extension to the temporal puzzle-solving paradigm.

\section{Our Approach}


Here, we present JPDVT, which uses a novel perspective on representing video and image data as non-ordered sets, to solve masked jigsaw puzzles. This representation enables JPDVT  to harness the inherent properties of  vision transformer architectures and the capabilities of conditional generative diffusion models to solve puzzles with missing pieces, as depicted in Fig.~\ref{fig_arch}.

\subsection{Masked Jigsaw Transformer Diffusion}


Consider an image or video, denoted as $\mathbf{X}$, partitioned into an unordered puzzle comprising individual pieces ${\mathbf{x}_1, \dots, \mathbf{x}_N}$ such that $\mathbf{x}_i \cap \mathbf{x}_j = \phi$ for all $i$ and $j$, and $\mathbf{X} = \cup_{i=1}^N \mathbf{x}_i$. Our objective is to reconstruct $\mathbf{X}$ by establishing a mapping $f(\mathbf{x}_i) = l_i$ that assigns the correct position to each piece. Specifically, we seek a solution where $\mathbf{X}$ is the union\footnote{In the cases with missing pieces or with eroded pieces, we will allow $\mathbf{X} \supseteq \cup_{i=1}^{M\le N} (\mathbf{x}_i @ l_i)$.} of the pieces placed at their designated locations, expressed as $\mathbf{X} = \cup_{i=1}^N (\mathbf{x}_i @ l_i)$, where $\mathbf{x}_i @ l_i$ signifies that piece $\mathbf{x}_i$ should be positioned at location $l_i$.


To achieve this goal, we propose representing the solved puzzle of $\mathbf{X}$ as a set of pairs denoted by $\mathbf{P} = \{(\mathbf{e}_1,\gamma(l_1)), \dots, (\mathbf{e}_N,\gamma(l_N))\}$. In this representation, $\mathbf{e}_i$ is an embedding of the visual content of piece $\mathbf{x}_i$, and $\gamma(l_i)$ is a vector containing the positional encoding of its location $l_i$, as illustrated in Fig.~\ref{fig_pe}. This representation is similar to the ones  used by masked autoencoders (MAE) to learn to inpaint missing data \cite{he2022masked}. However, a notable distinction lies in the fact that masked autoencoders possess knowledge of the locations of the given pieces, allowing them to directly work with the pairs $(\mathbf{e}_i,\gamma(l_i))$. In our scenario, where the pieces have been shuffled, a separate treatment of the visual content $\mathbf{e}_i$ and its corresponding location $\gamma(l_i)$ is necessary.

{\bf Puzzles without Missing Pieces.} Given a complete set of  puzzle pieces embeddings, $\textbf{E}= \{\mathbf{e}_1, \mathbf{e}_2, ..., \mathbf{e}_N\}$, let $\mathbf{L} = (\gamma(l_1), \gamma(l_2),\dots,\gamma(l_N))^T$ be a matrix with the sought positional encodings of the given pieces. Then, we  formulate solving the jigsaw puzzle as a conditional generation problem, where the goal is to learn the joint probability of the locations $\mathbf{L}$, given their unsorted visual content  $\mathbf{E}$:
$\mathbf{L}|\mathbf{E} \sim q\left( \mathbf{L} | \mathbf{E}\right )$.
We propose to learn this probability by using a  CDDPM as follows. 

For the  forward process, we diffuse the positional encodings, conditioned on the set of unsorted visual contents:
$$q(\mathbf{L}_t|\mathbf{L}_{t-1},\mathbf{E})=\mathcal{N}(\mathbf{L}_t|\mathbf{E};\sqrt{1 -\beta_t}\mathbf{L}_{t-1}|\mathbf{E},\beta_t\mathbf{I})$$
where noise is gradually added to the positional encodings of the patches (See  Fig.~\ref{fig_pe}),  and $\beta_t$ schedules the amount of noise added at each step $t$. The distribution of $\mathbf{L}_t$ given $\mathbf{L}_0$ can be explicitly expressed  by
$$q(\mathbf{L}_t|\mathbf{L}_0,\mathbf{E})=\mathcal{N}(\mathbf{L}_t|\mathbf{E};\sqrt{\alpha_t}\mathbf{L}_0|\mathbf{E},(1-\alpha_t)\mathbf{I})$$
where $\alpha_t = \prod_{s=1}^t(1 - \beta_s)$.

The reverse diffusion denoising process is given by:
$$p_{\theta}(\mathbf{L}_{t-1}|\mathbf{L}_t,\mathbf{E}):=\mathcal{N}(\textbf{L}_{t-1}|t,\mathbf{E};\mu_\theta(\textbf{L}_t, t, \mathbf{E}),\sigma_\theta(\textbf{L}_t , t, \mathbf{E})\textbf{I})$$ 
The conditional diffusion model learns a network $\epsilon_{\theta}$ to predict the noise added to the noisy input $\mathbf{L}_t$ with the loss: 
$$
     \mathcal{L}(\theta)=\mathbb{E}_{\textbf{L}_0\sim q(\textbf{L}_0),\epsilon \sim \mathcal{N}(0,I),t,\mathbf{E}} [\lVert \epsilon-\epsilon_\theta(\textbf{L}_t,t,\mathbf{E})\rVert ^2_2]
$$
where $ \textbf{L}_t=\sqrt{\alpha_t}\mathbf{L}_0 +\sqrt{(1-\alpha_t)}\epsilon$ for $\epsilon \sim {\cal{N}}(0,\mathbf{I})$.

{\bf Puzzles with  Missing Pieces.} For jigsaw puzzles with missing pieces, $\mathbf{E} = \mathbf{E}^g \cup \mathbf{E}^m$, where $\mathbf{E}^g$ and $\mathbf{E}^m$ are the given and missing pieces, respectively. In this case, we generate the positional information for both the given pieces and the missing ones. For the missing pieces, it is possible to fill them up using an inpainting algorithm \cite{suvorov2022resolution} after having their location, or alternatively, to generate the embeddings of the missing visual content simultaneously with the locations. In the later case, the conditional forward process  gradually adds noise to both, the positional encodings and the visual content embeddings of the missing pieces, conditioned on the visual content of the given pieces using
$q(\mathbf{L}_t,\mathbf{E}^m_t|\mathbf{L}_0,\mathbf{E}_0^m,\mathbf{E}^g)$
and the corresponding reverse diffusion process:
$p_{\theta}(\mathbf{L}_{t-1},\mathbf{E}_{t-1}^m|\mathbf{L}_t, \mathbf{E}_t^m,\mathbf{E}^g)$.



\begin{figure*}[t]
\centering
\includegraphics[width=1\textwidth]{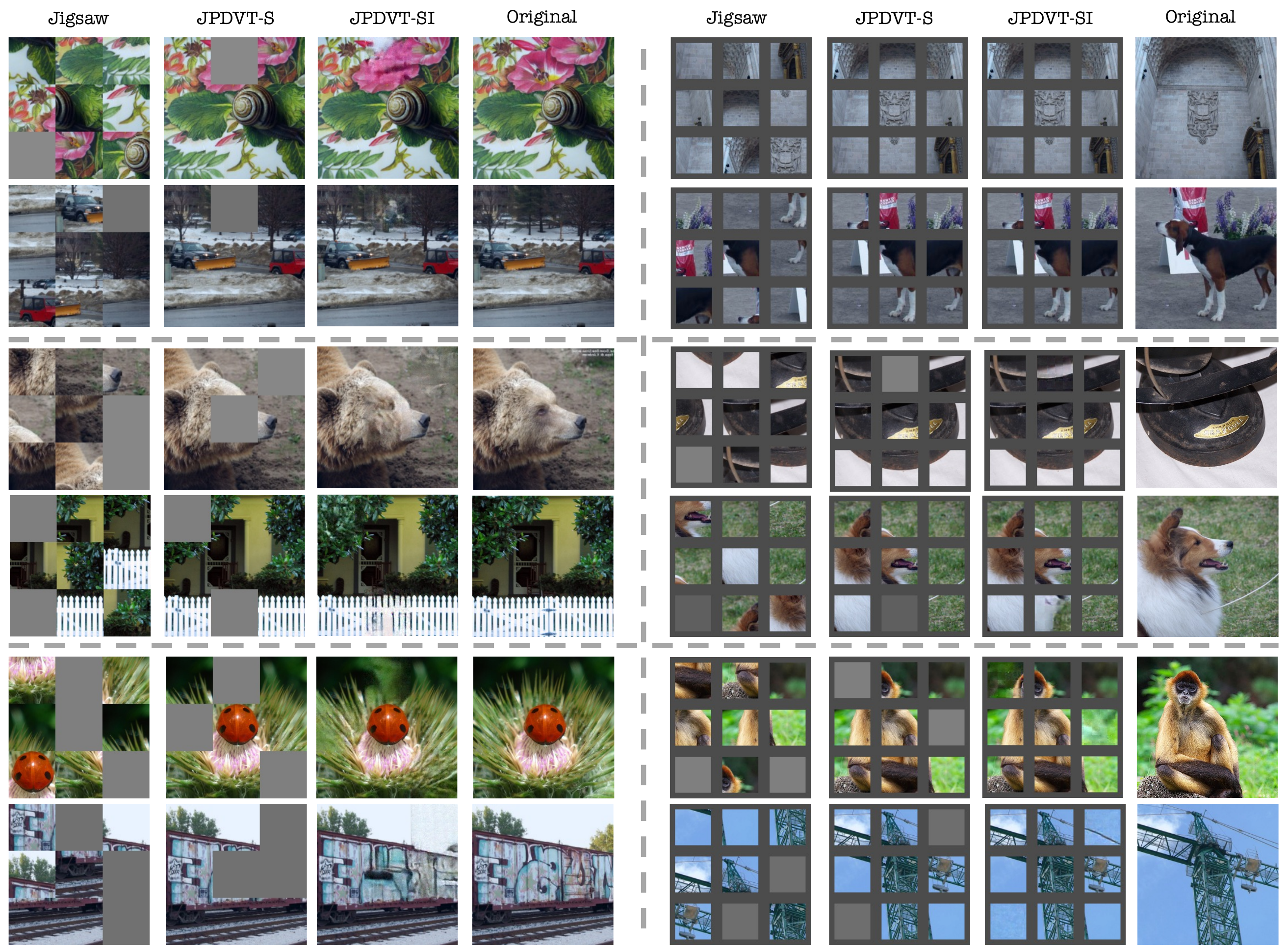}
\caption{Qualitative results for puzzles for Imagenet-1K dataset with 1, 2, and 3 missing pieces. Left: without  cropping. Right: with cropping. First column shows the input puzzle, second column shows the sorted result (denoted as JPDVT-S), third column shows the sorted and inpainted result (denoted as JPDVT-SI), and last column shows the ground truth images.}\label{fig:qual_imgs}
\end{figure*}

\begin{figure*}[tb]
\centering
\includegraphics[width=1\textwidth]{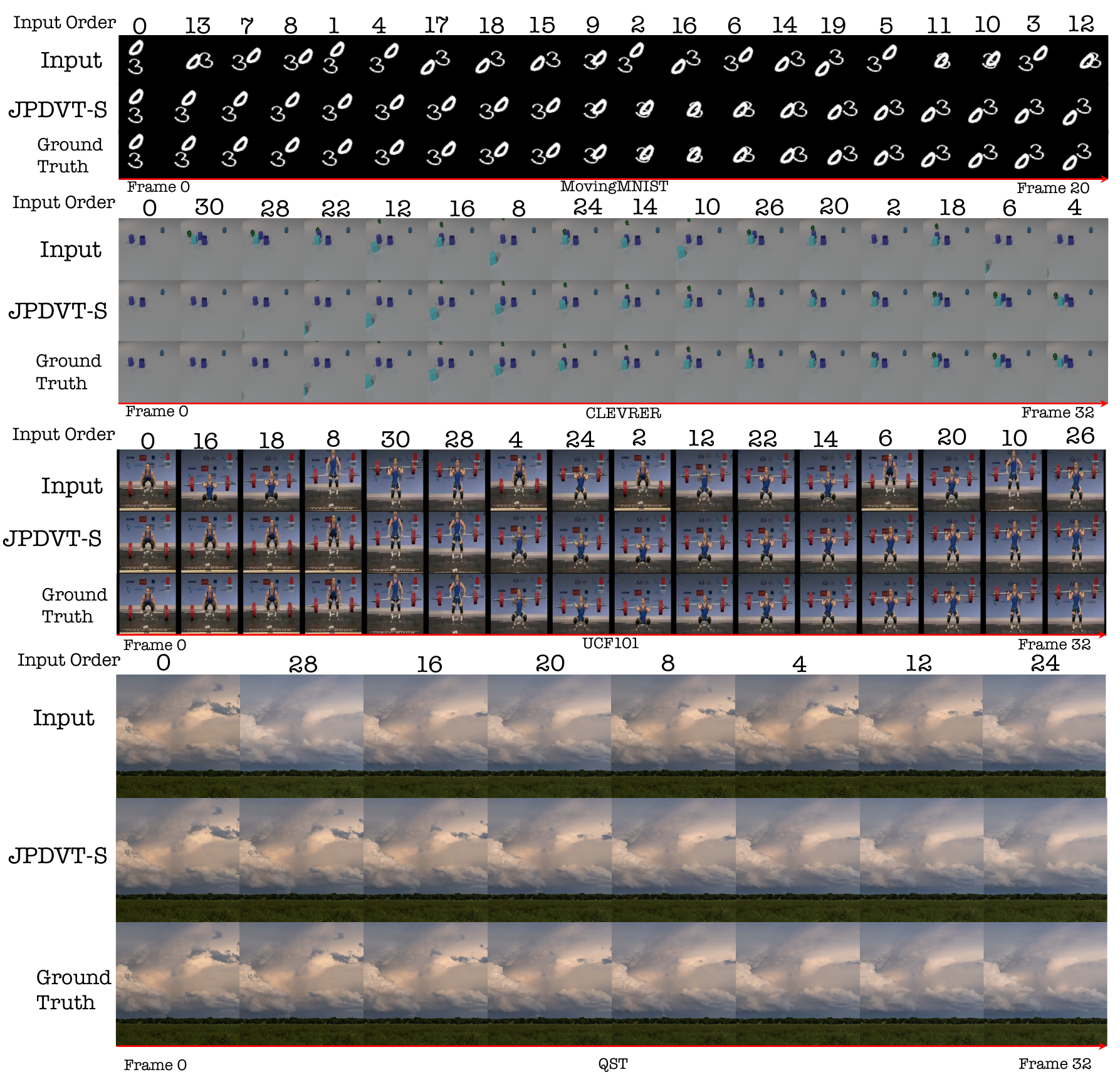}
\caption{Video reshuffle results are showcased on various datasets, including 20 frames on \textit{MovingMNIST}, 32 frames on \textit{CLEVRER}, 32 frames on \textit{UCF101}, and 32 frames on \textit{QST}. Due to space limitations, only a subset of frames is presented in this illustration. }\label{fig:video}
\end{figure*}

\begin{figure*}[tb]
\centering
\includegraphics[width=1\textwidth]{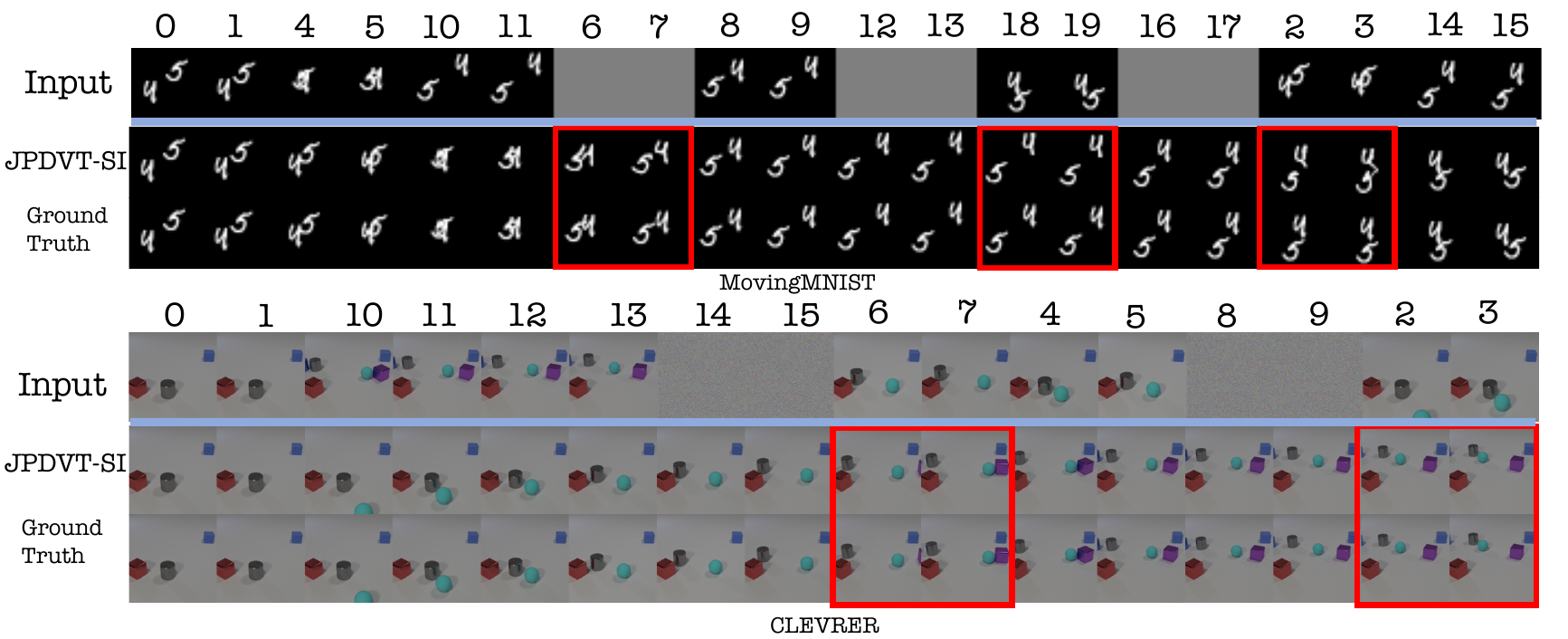}
\caption{Video reshuffle results with masks: Our diffusion model adeptly captures and reconstructs crucial latent components essential for the sequence reordering process. }\label{fig:video_masked}
\end{figure*}

\subsection{Implementation Details}

In this section, we give implementation details of our models, which are
 rooted in the official codebase of diffusion models from \cite{nichol2021improved}. Code will be provided. 

 \textbf{Solving image jigsaw puzzles. }
 We implemented our diffusion transformers by modifying ViTs based on \cite{beyer2022better}, following the approach in \cite{peebles2023scalable} for incorporating timesteps. Adaptive normalization layers are introduced before and after the multi-head attention and MLP modules to accommodate timesteps. To avoid the model reducing to a mere classification task, we anchor the positional embedding of the initial patch and represent subsequent patches using relative positional embeddings. This approach enhances the model's ability to capture spatial dependencies and ensures resilience against overreliance on absolute positional information. We utilized a linear noise schedule with $T=1000$ diffusion steps.

\textbf{Solving temporal jigsaw puzzles.} For low-resolution videos, with missing frames, we  applied diffusion  directly on the video pixels. For high-resolution videos, we followed \cite{rombach2022high} and first compressed each of the frames into a latent space, and subsequently  applied diffusion process on these latent variables.
For the reverse diffusion process, we modified the video U-Net based on \cite{singer2022make}. In our implementation, we only use the 2D spatial convolutional layers without the temporal convolutional layer, and in the  2D+1D attention layers, we removed the temporal positional embedding. We used a linear noise schedule, and we fixed diffusion steps to $T=1000$.

{ {\bf Positional Encoding.} We use the positional encoding:
$$ \gamma_{1D}(l_n) = 
\begin{cases}
    \gamma(l_n,2i) = \sin{(l_n/1000^{(2i/16)})}\\
    
    \gamma(l_n,2i+1) = \cos{(l_n/1000^{(2i/16)})}
  \end{cases}
$$
For temporal puzzles we use  a 1-D positional encoding with a dimension of 16  to represent each video frame. To align with the input feature map's shape, the positional encoding undergoes an initial transformation through an MLP, ensuring it matches the size of the input feature map. Subsequently, the positional encoding is concatenated with the feature maps.
For image puzzles, we employ a 2-D positional encoding with 32 dimensions to encode the position of an image patch, with the first half representing the x-coordinate and the second half representing the y-coordinate: $\gamma_{2D}(l_x, l_y) = [\gamma_{1D}(l_x), \gamma_{1D}(l_y)]$. This positional encoding is incorporated into image tokens following an MLP layer.

\section{Experiments}
Here, we describe the experiments we ran to validate the effectiveness of the proposed approach.
\subsection{Datasets}

We evaluated our approach on five datasets, consisting of both images and videos with different levels of resolution and complexity:

\noindent \textbf{\textit{ImageNet-1k}} \cite{deng2009imagenet} contains
1,281,167 training images, and 50,000 validation images categorized into 1000 classes.

\noindent \textbf{\textit{JPwLEG-3}} \cite{song2023siamese} contains 12,000 images collected from MET \cite{ypsilantis2021met} dataset. Each image is segmented into 9 fragments of dimensions 96 × 96, with gaps of 48 pixels between them. 9,000 images were allocated for training, 2,000 for testing, and an additional 1,000 for validation purposes.

\noindent \textbf{\textit{MovingMNIST}} \cite{srivastava2015unsupervised} comprises 10,000 video sequences, each containing 20 frames. Two digits move independently within each video, which has low resolution of $64 \times64$ pixels. We used 9,000 videos for training and tested our approach on the remaining 1,000 videos. 

\noindent \textbf{\textit{CLEVRER}} \cite{yi2019clevrer} is a diagnostic video dataset with 20,000 videos that simulate collisions among objects of different colors and shapes with $ 480 \times 320$ pixels.  The dataset is split into 10,000 videos for training, 5,000 for validation, and 5,000 for testing.

\noindent \textbf{\textit{QST}}(Quick Sky Time) \cite{zhang2020dtvnet} contains 1,167 video clips extracted from 216 time-lapse 4K videos collected from YouTube, totaling 285,446 frames. The resolution of each frame exceeds $1,024\times 1,024$. We downsample the frames to $256 \time 256$, and train on 1,000 videos, test on 167 videos.

\noindent \textbf{\textit{UCF101}} \cite{soomro2012ucf101} is an action recognition dataset, consisting of 13,320 video clips representing 101 different human actions. Each frame has a resolution of $320 \times 240$ pixels, with 9,537 videos  for training and 3,783 videos for testing.



\subsection{Metrics}


\textbf{\textit{Image Puzzle Solving Accuracy}}: 
In assessing image jigsaw puzzle solutions, we gauge solving accuracy both at the puzzle and piece levels. The puzzle-wise accuracy metric denotes the percentage of flawlessly reassembled image puzzles—essentially, the fraction of puzzles with every single piece correctly positioned, totaling 100\%. On the other hand, the piece-wise accuracy metric quantifies the ratio of correctly placed pieces to the overall number of pieces in the dataset. This comprehensive evaluation framework provides a nuanced understanding of the reconstruction quality, capturing both holistic puzzle-level performance and the precision of individual pieces within the puzzle assembly.

\noindent \textbf{\textit{Video Puzzle Solving Accuracy}} Following \cite{dicle2016solving}, we measure video jigsaw puzzle solving accuracy with the  average \textit{Normalized Kendall distance}, based on 1,000 testing samples from each dataset. The \textit{Kendall distance}, also called \textit{bubble-sort distance}, scores the number of pairwise disagreements between the candidate sorting and the ground truth, which is defined as the number of elements in the set constructed from two sequences $\sigma_1$ and $\sigma_2$ as follows: 
  $$ \mathcal{G}(\sigma_1,\sigma_2) = \{(i,j)|\sigma_1(i)<\sigma_2(j),\sigma_2(i)> \sigma_1(j)\}$$
where $\sigma_1(i)$ and $\sigma_2(i)$ are the rankings of the element $i$ in the $\sigma_1$ and $\sigma_2$ sequences. The normalized distance is the Kendall distance, normalized by the biggest possible distance. The lower the distance is, the more similar the two sequences are, with the worst possible score being 1. 


\subsection{Image Jigsaw Puzzles Experiments}

We conducted experiments on solving image jigsaw puzzles using the \textit{ImangeNet-1k} dataset, \textit{JPwLEG-3} \cite{song2023siamese}, and the dataset used in \cite{bridger2020solving}. 
Our models were trained separately with varying hyperparameters tailored to accommodate different numbers of puzzle pieces. 
To assess JPDVT's performance, we conducted comparative evaluations against state-of-the-art methods \cite{wei2019iterative, song2023siamese, bridger2020solving, paumard2020deepzzle, song2023solving}. 


The experiments on \textit{ImangeNet-1k} follow the data preprocessing procedure from \cite{wei2019iterative}. During training, each image underwent an initial resizing to $255 \times 255$ before being divided into nine patches, each of size $85 \times 85$. Subsequently, every patch was randomly cropped to $64 \times 64$. A total of $9!=362,880$ permutations were applied to these patches, and the models were trained to generate permutation indices based on the image patches. Data augmentation solely involved horizontal flipping. During testing, in contrast to random cropping, we used the cropped patch from the center. The experiments on \textit{JPwLEG-3 } follow the experiment setting in \cite{song2023siamese}. In the experiments with a large number of pieces, we follow the experiment setting with the largest number of pieces in \cite{bridger2020solving}.

Fig.~\ref{fig:qual_imgs} presents qualitative results for puzzles with 1, 2, and 3 missing pieces, showcasing flawless reshuffling. Quantitative results for puzzles without missing pieces are provided in  tables~\ref{tab:imagepuzzle}, ~\ref{tab:largepiecesnumber} and  ~\ref{tab:otherSOTA}. JPDVT achieved a 68.7\% success rate on \textit{imagenet-1k} dataset and 71.3\% on \textit{JPwLEG-3} dataset for puzzles without missing pieces—an improvement of 20.7\% and 11.6\%, respectively, over prior state-of-the-art performances at the puzzle level. 
The results in table~\ref{tab:largepiecesnumber} show the capability of our model to handle a large number of puzzle pieces.
Unlike previous methods, our approach handles puzzles with missing fragments, as demonstrated in Table~\ref{tab:imagepuzzle_masked}. The performance gradually decreases with an increasing number of random missing pieces, indicating greater difficulty with eroded pieces compared to those without erosion. This suggests the architecture effectively learns to align piece boundaries, a crucial aspect in solving the puzzle.



\begin{table}[t]
\centering
\caption{
Results of $3 \times 3$ image puzzle solving on \textit{ImageNet-1k} val split. Puzzle-level shows the percentages of flawlessly reassembled puzzles. Fragment-level shows the fraction of correctly placed pieces of all the pieces in the dataset.
}
\vspace{-0.15in}
\begin{tabular}{*3c}
Method & Puzzle-level & Piece-level     \\ 
\hline
SAJ\cite{wei2019iterative} & 47.3\%  &  NA\\
Deepzzle \cite{paumard2020deepzzle}& 48\% & 78\%\\
JPDVT(Ours) & \cellcolor{gray!25}  \textbf{68.7\%} &\cellcolor{gray!25}  \textbf{83.3}\%
\end{tabular} 

\vspace{-0.1in}
\label{tab:imagepuzzle}
\end{table}

\begin{table}[t]
\centering
\caption{
Results on $3 \times 3$ puzzle recognition on \textit{ImageNet-1K val} split on our proposed model on different number of missing fragments with and without eroded gaps.
}\label{tbl:video}
\vspace{-0.15in}
\begin{tabular}{*5c}
\% Missing & \multicolumn{2}{c}{With gap}  & \multicolumn{2}{c}{Without gap}    \\ 
\hline
& \small puzzle & \small piece & \small puzzle &  \small piece \\
\cline{2-5}
0 & \cellcolor{gray!25}\textbf{68.7}\%&\cellcolor{gray!25} \textbf{83.3}\%& \cellcolor{gray!25} \textbf{99.4}\%  &\cellcolor{gray!25}\textbf{99.6}\% \\
11.1 & 41.5\% & 72.0\%& 96.9\% &98.8\%\\
22.2 & 21.4\% & 61.8\% & 82.8\%& 95.0\% \\
33.3 &14.9 \% & 54.1 \% & 56.1\%& 76.4\%\\
\end{tabular} 

\vspace{-0.1in}
\label{tab:imagepuzzle_masked}
\end{table}

\begin{table}[t]
\centering
\caption{
Results of 150 pieces puzzle solving with $7\%$  erosion. 
}

\vspace{-0.3cm}
\begin{tabular}{*3c}

Method & Piece-level&  Puzzle-level    \\ 
\hline
SJP-ED\cite{bridger2020solving} & 66.7\%  &  10\%\\
JPDVT(Ours) &  \textbf{75.9}\%  &  \textbf{45\%}
\end{tabular} 
\vspace{-0.15in}
\label{tab:largepiecesnumber}
\end{table}



\vspace{-0.25cm}
\begin{table}[t]
\centering
\caption{
Comparison to the SOTA of $3 \times 3$ puzzle recognition on the JPwLEG-3 \cite{song2023siamese} dataset.
}
\vspace{-0.3cm}
\begin{tabular}{*3c}

Method & Venue & Piece-level    \\ 
\hline
Siamese \cite{song2023siamese} & AAAI-2023  & 59.7\% \\
Puzzlet \cite{song2023solving} & ICASSP-2023 & 58.2\% \\
SJP-ED \cite{bridger2020solving}& CVPR-2020&  55.2\% \\
Deepzzle \cite{paumard2020deepzzle} & TIP-2020& 52.3\%\\
JPDVT(Ours) & -&  \textbf{71.3\% } \\
\end{tabular} 
\vspace{-0.1in}
\label{tab:otherSOTA}
\end{table}

\subsection{Temporal Jigsaw Puzzles  Experiments}

\begin{table}
\caption{Normalized Kendall Distance for Temporal Puzzles without missing data measures the number of pairwise disagreements between two sequences.}
\vspace{-0.5cm}
\begin{center}
\begin{adjustbox}{width=\linewidth}
\begin{tabular}{*6c}
Datasets & Method &\multicolumn{4}{l} {Normalized Kendall Distance x $10^3$ } \\
\toprule
Size, \#Pieces & & 1, 20& 2, 10 & 4, 5 & \\
 \cline{3-5}
\multirow{4}{*}{MMNIST}&  STP& 1.4 & 0.04 & 0.01\\ 
&  VCOP& - & - & 13.7 \\
& J-VAD& {173} & 0.46& \textbf{0}\\
& \bf{JPDVT(Ours)}&  \cellcolor{gray!25}\bf{0.7 }& \cellcolor{gray!25}\bf{0.04} &\cellcolor{gray!25} \textbf{0}\\
\hline
Size, \#Pieces & & 1, 32& 2, 16 & 4, 8 & 8, 4  \\
\cline{3-6}
\multirow{4}{*}{CLEVRER} & STP & - &0.15 & 0 & 0\\
&VCOP&- & -& -& 4.6 \\
&J-VAD & {268.2}&58.2 & 0.04 & 0\\
& \bf{JPDVT(Ours)}& \cellcolor{gray!25}\bf{2.6} & \cellcolor{gray!25}\bf{0.17} &\cellcolor{gray!25} \bf{0.04}& \cellcolor{gray!25}\bf{0}\\
\hline
\multirow{4}{*}{UCF} 
&VCOP&- & -& -& 14.5 \\
&J-VAD & -&{164.3} & 3.8 & 0.5\\
& \bf{JPDVT(Ours)}& \cellcolor{gray!25}\bf{26.2} & \cellcolor{gray!25}\bf{4.2} & \cellcolor{gray!25}\bf{1.2}& \cellcolor{gray!25}\bf{0.3}\\
\hline
\multirow{4}{*}{QST} & STP & - & 0 &0 & 0\\
&VCOP&- & -& -& 14.5 \\
& \bf{JPDVT(Ours)}& \cellcolor{gray!25}{7.0} &\cellcolor{gray!25} {3.0} & \cellcolor{gray!25}{0.19}& \cellcolor{gray!25}{0.04}\\

\end{tabular}
\end{adjustbox}
\vspace{-.8cm}
\end{center}
\end{table}

We constructed temporal puzzles from videos by segmenting them,  where each piece is a clip comprising one or more frames  in the correct temporal order. Specifically, in the case of MovingMNIST, we built puzzles with: 20 pieces, where each piece is a single frame; 10 pieces, each comprising a pair of consecutive frames; and 5 pieces, with each piece containing four consecutive frames. Similarly, for the CLEVRER, UCF, and QST datasets, our evaluation included puzzles with 32 pieces (single frames), 16 pieces (two frames), 8 pieces (four frames), and 4 pieces (eight frames). Qualitative examples in Fig.~\ref{fig:video} and \ref{fig:video_masked} demonstrate instances where our approach adeptly resolved the puzzles  without and with missing data, respectively.

 We compared the performance of our algorithm  solving temporal puzzles, without missing data, against the state-of-the-art methods STP (Space-Time Puzzle) \cite{dicle2016solving}, VCOP (Video Clip-Order Prediction) \cite{xu2019self}, and J-VAD (Joint Video Anomaly Detection) \cite{wang2022video}, using four video datasets, whenever possible, as described below. We trained these models following the processes detailed in their respective papers, with adjustments made to network sizes to accommodate varying dataset dimensions. To ensure fair comparisons, we fixed the first frame as an anchor frame, and fully randomly shuffle the remaining frames with no mask. Detailed training hyperparameters are available in the appendix.

 STP  adapts its inputs based on video characteristics. For dynamic textures, exemplified by the QST dataset, it utilizes the PCA decomposition of pixels. However, for more structured videos, such as MovingMNIST and CLEVRER, STP requires  feature correspondences across the input frames. Thus, for MovingMNIST and CLEVRER we ran experiments by providing STP  ground-truth object coordinates, giving this method an unfair advantage. Regrettably, providing feature correspondences  for the UCF dataset was impractical due to its complexity. Additionally, the computational demands of STP rendered  unfeasible to run for puzzles with 32 pieces, since its inference time for this setting takes over 30 minutes/video. In contrast, our approach takes approximately 10 seconds/video.

 VCOP and J-VAD, on the other hand,  have faster inference times, approximately 1 sec/video. However, VCOP predicts the permutation of videos by classifying over a vast number of possibilities. For instance, with 10 clips, the model must classify among $10! = 3,628,800$ classes. As the number of video clips for ordering increases, training the model becomes impractical due to the exponential growth in classes. J-VAD, employs a 3D-CNN model coupled with a classifier to directly predict the position of each input frame or clip. However, its performance experiences a significant decline when tested on 16  and 32 clips. This phenomenon aligns with findings in \cite{wang2022video}, where performance deterioration is reported for frame numbers larger than 9, consistent with our observations.

Table~\ref{tbl:video} provides the quantitative results of the comparisons against the baselines.  Our approach achieves State-Of-The-Art performance across the board and second best for the QST dataset. While STP has comparable performance in some settings and best performance for QST, as noted above, it uses different types of inputs depending on the video characteristics and in this case ground-truth feature correspondences, while our approach always works directly on the raw data. 
In addition, while our approach is an order of magnitude slower than VCOP and J-VAD, it  can handle all the testing settings while the others cannot. 

Finally, JPDVT can handle temporal puzzles with missing fragments, as shown in Fig.~\ref{fig:video_masked}. As seen in Fig.~\ref{fig:video_plots}, JPDVT's performance gradually decreases with an increasing number of random missing pieces.  This capability  allows the use of JPDVT to increase the temporal resolution of videos, as illustrated in the supplemental material.

\begin{figure}[tb]
\centering
\begin{adjustbox}{width=\linewidth}
\begin{tabular}{cc}
\includegraphics[width=0.5\textwidth]{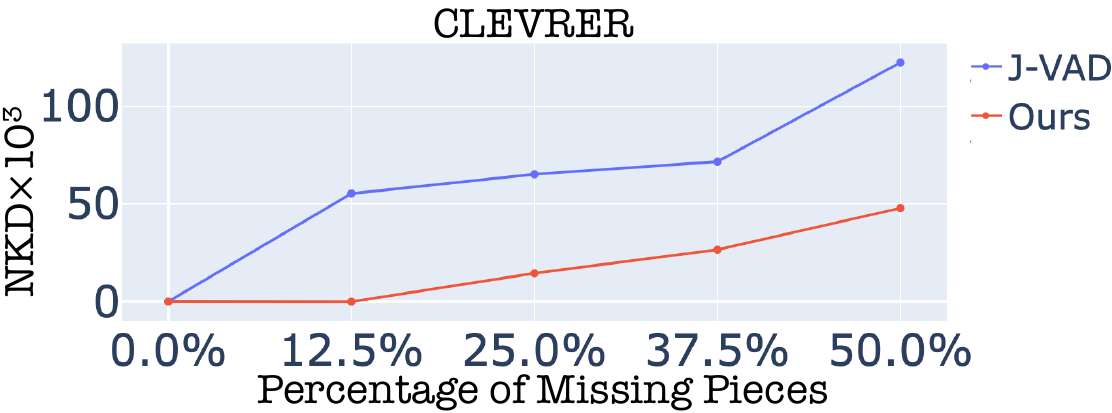}  & 
\includegraphics[width=0.5\textwidth]{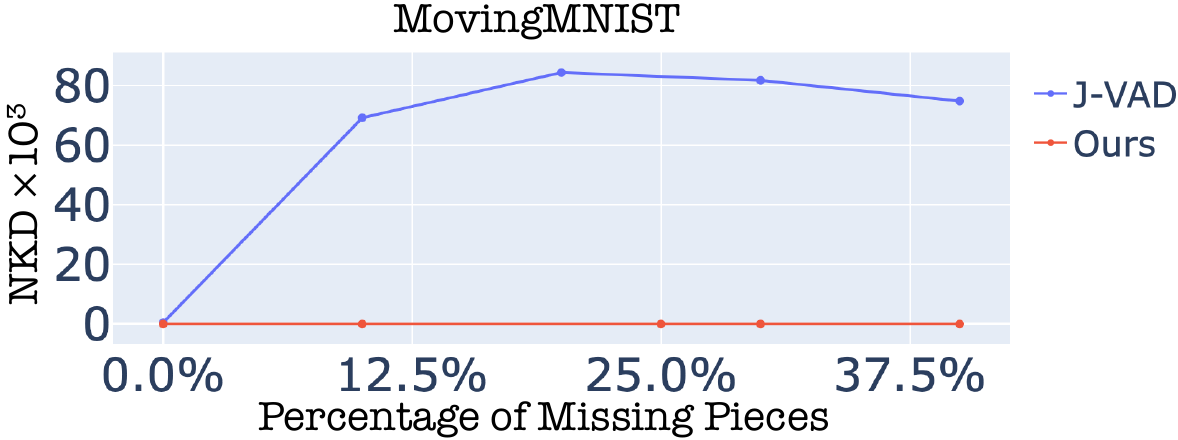}\\
\large (a) &
 \large(b) \\
 \includegraphics[width=0.5\textwidth]{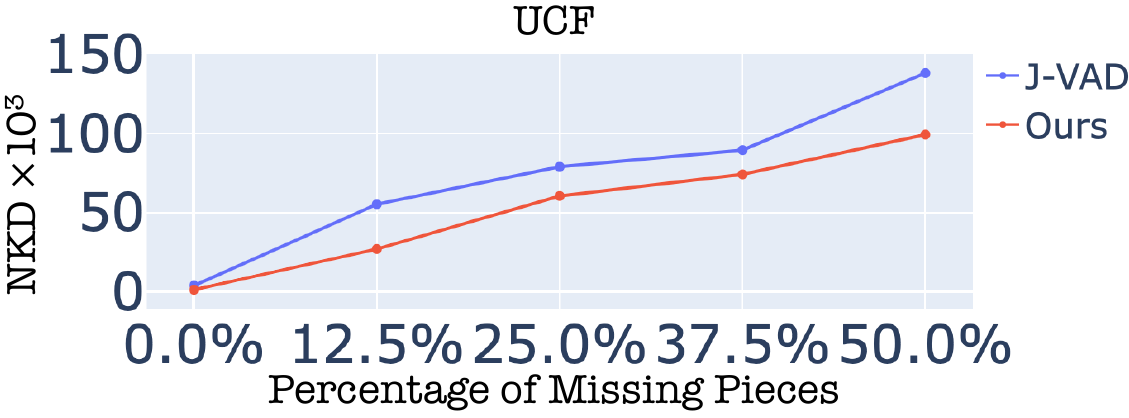} &
 \includegraphics[width=0.5\textwidth]{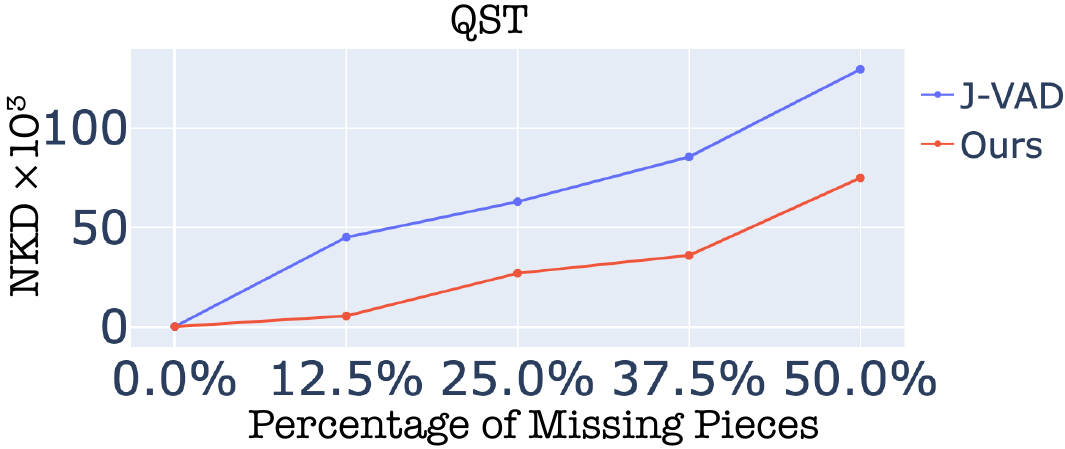}\\
\large(c) &
 \large(d) \\
 
 \end{tabular}
 \end{adjustbox}
 \vspace{-0.3cm}
 \caption{Normalized Kendall distances vs. missing pieces. (a) CLEVRER dataset with 8 pieces, each containing 4 frames. (b) MovingMNIST dataset with 8 pieces, each with 4 frames. (c) UCF dataset with 8 pieces, each with 4 frames.
 (d) QST dataset with 8 pieces, each with 4 frames.
 }\label{fig:video_plots}
 \vspace{-0.5cm}
 \end{figure}

\section{Conclusion}

Solving  puzzles is a fundamental task involving the reassembly of disordered fragments, a classic problem in   pattern recognition that traces its roots back to 1964 \cite{freeman1964apictorial}. 
In this paper, we present JPDVT, a novel method for solving image and temporal puzzles in an unified framework. Our approach works by describing the underlying data as an unordered set of pairs, associating the (2D or 1D) positional encoding of a puzzle piece with an embedding of its visual content. This representation enables the formulation of the puzzle-solving problem as a Conditional Diffusion Denoising process, where the unknown positional encodings of the shuffled puzzle pieces are recovered through a reverse denoising process, conditioned on the visual content embeddings of the given  elements. JPDVT   achieves state-of-the-art performance in quantitative  comparisons and produces high-quality qualitative examples, while solving  image and video puzzles. Notably, it can handle missing data and  solve  larger puzzles than previous approaches. 

{
    \small
    \bibliographystyle{ieeenat_fullname}
    \bibliography{main}

\begin{thebibliography}{36}
\providecommand{\natexlab}[1]{#1}
\providecommand{\url}[1]{\texttt{#1}}
\expandafter\ifx\csname urlstyle\endcsname\relax
  \providecommand{\doi}[1]{doi: #1}\else
  \providecommand{\doi}{doi: \begingroup \urlstyle{rm}\Url}\fi

\bibitem[Basha et~al.(2012)Basha, Moses, and Avidan]{basha2012photo}
Tali Basha, Yael Moses, and Shai Avidan.
\newblock Photo sequencing.
\newblock In \emph{Computer Vision--ECCV 2012: 12th European Conference on Computer Vision, Florence, Italy, October 7-13, 2012, Proceedings, Part VI 12}, pages 654--667. Springer, 2012.

\bibitem[Beyer et~al.(2022)Beyer, Zhai, and Kolesnikov]{beyer2022better}
Lucas Beyer, Xiaohua Zhai, and Alexander Kolesnikov.
\newblock Better plain vit baselines for imagenet-1k.
\newblock \emph{arXiv preprint arXiv:2205.01580}, 2022.

\bibitem[Bridger et~al.(2020)Bridger, Danon, and Tal]{bridger2020solving}
Dov Bridger, Dov Danon, and Ayellet Tal.
\newblock Solving jigsaw puzzles with eroded boundaries.
\newblock In \emph{Proceedings of the IEEE/CVF Conference on Computer Vision and Pattern Recognition}, pages 3526--3535, 2020.

\bibitem[Dekel et~al.(2014)Dekel, Moses, and Avidan]{dekel2014photo}
Tali Dekel, Yael Moses, and Shai Avidan.
\newblock Photo sequencing.
\newblock \emph{International journal of computer vision}, 110:\penalty0 275--289, 2014.

\bibitem[Deng et~al.(2009)Deng, Dong, Socher, Li, Li, and Fei-Fei]{deng2009imagenet}
Jia Deng, Wei Dong, Richard Socher, Li-Jia Li, Kai Li, and Li Fei-Fei.
\newblock Imagenet: A large-scale hierarchical image database.
\newblock In \emph{2009 IEEE conference on computer vision and pattern recognition}, pages 248--255. Ieee, 2009.

\bibitem[Dicle et~al.(2016)Dicle, Yilmaz, Camps, and Sznaier]{dicle2016solving}
Caglayan Dicle, Burak Yilmaz, Octavia Camps, and Mario Sznaier.
\newblock Solving temporal puzzles.
\newblock In \emph{Proceedings of the IEEE Conference on Computer Vision and Pattern Recognition}, pages 5896--5905, 2016.

\bibitem[Dosovitskiy et~al.(2021)Dosovitskiy, Beyer, Kolesnikov, Weissenborn, Zhai, Unterthiner, Dehghani, Minderer, Heigold, Gelly, Uszkoreit, and Houlsby]{dosovitskiy2021an}
Alexey Dosovitskiy, Lucas Beyer, Alexander Kolesnikov, Dirk Weissenborn, Xiaohua Zhai, Thomas Unterthiner, Mostafa Dehghani, Matthias Minderer, Georg Heigold, Sylvain Gelly, Jakob Uszkoreit, and Neil Houlsby.
\newblock An image is worth 16x16 words: Transformers for image recognition at scale.
\newblock In \emph{International Conference on Learning Representations}, 2021.

\bibitem[Freeman and Garder(1964)]{freeman1964apictorial}
Herbert Freeman and L Garder.
\newblock Apictorial jigsaw puzzles: The computer solution of a problem in pattern recognition.
\newblock \emph{IEEE Transactions on Electronic Computers}, \penalty0 (2):\penalty0 118--127, 1964.

\bibitem[He et~al.(2022)He, Chen, Xie, Li, Doll{\'a}r, and Girshick]{he2022masked}
Kaiming He, Xinlei Chen, Saining Xie, Yanghao Li, Piotr Doll{\'a}r, and Ross Girshick.
\newblock Masked autoencoders are scalable vision learners.
\newblock In \emph{Proceedings of the IEEE/CVF conference on computer vision and pattern recognition}, pages 16000--16009, 2022.

\bibitem[Ho et~al.(2020)Ho, Jain, and Abbeel]{ho2020denoising}
Jonathan Ho, Ajay Jain, and Pieter Abbeel.
\newblock Denoising diffusion probabilistic models.
\newblock \emph{Advances in neural information processing systems}, 33:\penalty0 6840--6851, 2020.

\bibitem[Kim et~al.(2019)Kim, Cho, and Kweon]{kim2019self}
Dahun Kim, Donghyeon Cho, and In~So Kweon.
\newblock Self-supervised video representation learning with space-time cubic puzzles.
\newblock In \emph{Proceedings of the AAAI conference on artificial intelligence}, pages 8545--8552, 2019.

\bibitem[Kim et~al.(2021)Kim, Wu, Dai, Zhang, Yan, Vajda, and Kim]{kim2021rethinking}
Kyungmin Kim, Bichen Wu, Xiaoliang Dai, Peizhao Zhang, Zhicheng Yan, Peter Vajda, and Seon~Joo Kim.
\newblock Rethinking the self-attention in vision transformers.
\newblock In \emph{Proceedings of the IEEE/CVF Conference on Computer Vision and Pattern Recognition}, pages 3071--3075, 2021.

\bibitem[Lee et~al.(2017)Lee, Huang, Singh, and Yang]{lee2017unsupervised}
Hsin-Ying Lee, Jia-Bin Huang, Maneesh Singh, and Ming-Hsuan Yang.
\newblock Unsupervised representation learning by sorting sequences.
\newblock In \emph{Proceedings of the IEEE international conference on computer vision}, pages 667--676, 2017.

\bibitem[Li et~al.(2021)Li, Liu, Wang, Liu, and Zeng]{li2021jigsawgan}
Ru Li, Shuaicheng Liu, Guangfu Wang, Guanghui Liu, and Bing Zeng.
\newblock Jigsawgan: Auxiliary learning for solving jigsaw puzzles with generative adversarial networks.
\newblock \emph{IEEE Transactions on Image Processing}, 31:\penalty0 513--524, 2021.

\bibitem[Misra et~al.(2016)Misra, Zitnick, and Hebert]{misra2016shuffle}
Ishan Misra, C~Lawrence Zitnick, and Martial Hebert.
\newblock Shuffle and learn: unsupervised learning using temporal order verification.
\newblock In \emph{Computer Vision--ECCV 2016: 14th European Conference, Amsterdam, The Netherlands, October 11--14, 2016, Proceedings, Part I 14}, pages 527--544. Springer, 2016.

\bibitem[Moses et~al.(2013)Moses, Avidan, et~al.]{moses2013space}
Yael Moses, Shai Avidan, et~al.
\newblock Space-time tradeoffs in photo sequencing.
\newblock In \emph{Proceedings of the IEEE International Conference on Computer Vision}, pages 977--984, 2013.

\bibitem[Nichol and Dhariwal(2021)]{nichol2021improved}
Alexander~Quinn Nichol and Prafulla Dhariwal.
\newblock Improved denoising diffusion probabilistic models.
\newblock In \emph{International Conference on Machine Learning}, pages 8162--8171. PMLR, 2021.

\bibitem[Noroozi and Favaro(2016)]{noroozi2016unsupervised}
Mehdi Noroozi and Paolo Favaro.
\newblock Unsupervised learning of visual representations by solving jigsaw puzzles.
\newblock In \emph{European conference on computer vision}, pages 69--84. Springer, 2016.

\bibitem[Paumard and \etal(2020)]{paumard2020deepzzle}
M. Paumard and \etal.
\newblock Deepzzle: Solving visual jigsaw puzzles with deep learning and shortest path optimization.
\newblock \emph{IEEE Trans. on Image Proc.}, 29:\penalty0 3569--3581, 2020.

\bibitem[Paumard et~al.(2018)Paumard, Picard, and Tabia]{paumard2018image}
Marie-Morgane Paumard, David Picard, and Hedi Tabia.
\newblock Image reassembly combining deep learning and shortest path problem.
\newblock In \emph{Proceedings of the European conference on computer vision (ECCV)}, pages 153--167, 2018.

\bibitem[Peebles and Xie(2023)]{peebles2023scalable}
William Peebles and Saining Xie.
\newblock Scalable diffusion models with transformers.
\newblock In \emph{Proceedings of the IEEE/CVF International Conference on Computer Vision}, pages 4195--4205, 2023.

\bibitem[Rombach et~al.(2022)Rombach, Blattmann, Lorenz, Esser, and Ommer]{rombach2022high}
Robin Rombach, Andreas Blattmann, Dominik Lorenz, Patrick Esser, and Bj{\"o}rn Ommer.
\newblock High-resolution image synthesis with latent diffusion models.
\newblock In \emph{Proceedings of the IEEE/CVF conference on computer vision and pattern recognition}, pages 10684--10695, 2022.

\bibitem[Singer et~al.(2022)Singer, Polyak, Hayes, Yin, An, Zhang, Hu, Yang, Ashual, Gafni, et~al.]{singer2022make}
Uriel Singer, Adam Polyak, Thomas Hayes, Xi Yin, Jie An, Songyang Zhang, Qiyuan Hu, Harry Yang, Oron Ashual, Oran Gafni, et~al.
\newblock Make-a-video: Text-to-video generation without text-video data.
\newblock In \emph{The Eleventh International Conference on Learning Representations}, 2022.

\bibitem[Song and \etal(2023{\natexlab{a}})]{song2023siamese}
X. Song and \etal.
\newblock Siamese-discriminant deep reinforcement learning for solving jigsaw puzzles with large eroded gaps.
\newblock In \emph{AAAI}, 2023{\natexlab{a}}.

\bibitem[Song and \etal(2023{\natexlab{b}})]{song2023solving}
X. Song and \etal.
\newblock Solving jigsaw puzzle of large eroded gaps using puzzlet discriminant network.
\newblock In \emph{ICASSP}, 2023{\natexlab{b}}.

\bibitem[Soomro et~al.(2012)Soomro, Zamir, and Shah]{soomro2012ucf101}
Khurram Soomro, Amir~Roshan Zamir, and Mubarak Shah.
\newblock Ucf101: A dataset of 101 human actions classes from videos in the wild.
\newblock \emph{arXiv preprint arXiv:1212.0402}, 2012.

\bibitem[Srivastava et~al.(2015)Srivastava, Mansimov, and Salakhudinov]{srivastava2015unsupervised}
Nitish Srivastava, Elman Mansimov, and Ruslan Salakhudinov.
\newblock Unsupervised learning of video representations using lstms.
\newblock In \emph{International conference on machine learning}, pages 843--852. PMLR, 2015.

\bibitem[Suvorov et~al.(2022)Suvorov, Logacheva, Mashikhin, Remizova, Ashukha, Silvestrov, Kong, Goka, Park, and Lempitsky]{suvorov2022resolution}
Roman Suvorov, Elizaveta Logacheva, Anton Mashikhin, Anastasia Remizova, Arsenii Ashukha, Aleksei Silvestrov, Naejin Kong, Harshith Goka, Kiwoong Park, and Victor Lempitsky.
\newblock Resolution-robust large mask inpainting with fourier convolutions.
\newblock In \emph{Proceedings of the IEEE/CVF winter conference on applications of computer vision}, pages 2149--2159, 2022.

\bibitem[Vaswani et~al.(2017)Vaswani, Shazeer, Parmar, Uszkoreit, Jones, Gomez, Kaiser, and Polosukhin]{vaswani2017attention}
Ashish Vaswani, Noam Shazeer, Niki Parmar, Jakob Uszkoreit, Llion Jones, Aidan~N Gomez, {\L}ukasz Kaiser, and Illia Polosukhin.
\newblock Attention is all you need.
\newblock \emph{Advances in neural information processing systems}, 30, 2017.

\bibitem[Wang et~al.(2022)Wang, Wang, Qin, Zhang, Bao, and Huang]{wang2022video}
Guodong Wang, Yunhong Wang, Jie Qin, Dongming Zhang, Xiuguo Bao, and Di Huang.
\newblock Video anomaly detection by solving decoupled spatio-temporal jigsaw puzzles.
\newblock In \emph{European Conference on Computer Vision}, pages 494--511. Springer, 2022.

\bibitem[Wei et~al.(2019)Wei, Xie, Ren, Xia, Su, Liu, Tian, and Yuille]{wei2019iterative}
Chen Wei, Lingxi Xie, Xutong Ren, Yingda Xia, Chi Su, Jiaying Liu, Qi Tian, and Alan~L Yuille.
\newblock Iterative reorganization with weak spatial constraints: Solving arbitrary jigsaw puzzles for unsupervised representation learning.
\newblock In \emph{Proceedings of the IEEE/CVF Conference on Computer Vision and Pattern Recognition}, pages 1910--1919, 2019.

\bibitem[Xiao et~al.(2023)Xiao, Fu, Zhou, Liu, and Zha]{xiao2023random}
Jie Xiao, Xueyang Fu, Man Zhou, Hongjian Liu, and Zheng-Jun Zha.
\newblock Random shuffle transformer for image restoration.
\newblock In \emph{International Conference on Machine Learning}, pages 38039--38058. PMLR, 2023.

\bibitem[Xu et~al.(2019)Xu, Xiao, Zhao, Shao, Xie, and Zhuang]{xu2019self}
Dejing Xu, Jun Xiao, Zhou Zhao, Jian Shao, Di Xie, and Yueting Zhuang.
\newblock Self-supervised spatiotemporal learning via video clip order prediction.
\newblock In \emph{Proceedings of the IEEE/CVF Conference on Computer Vision and Pattern Recognition}, pages 10334--10343, 2019.

\bibitem[Yi et~al.(2019)Yi, Gan, Li, Kohli, Wu, Torralba, and Tenenbaum]{yi2019clevrer}
Kexin Yi, Chuang Gan, Yunzhu Li, Pushmeet Kohli, Jiajun Wu, Antonio Torralba, and Joshua~B Tenenbaum.
\newblock Clevrer: Collision events for video representation and reasoning.
\newblock \emph{arXiv preprint arXiv:1910.01442}, 2019.

\bibitem[Ypsilantis et~al.(2021)Ypsilantis, Garcia, Han, Ibrahimi, Van~Noord, and Tolias]{ypsilantis2021met}
Nikolaos-Antonios Ypsilantis, Noa Garcia, Guangxing Han, Sarah Ibrahimi, Nanne Van~Noord, and Giorgos Tolias.
\newblock The met dataset: Instance-level recognition for artworks.
\newblock In \emph{Thirty-fifth Conference on Neural Information Processing Systems Datasets and Benchmarks Track (Round 2)}, 2021.

\bibitem[Zhang et~al.(2020)Zhang, Xu, Liu, Wang, Wu, Liu, and Jiang]{zhang2020dtvnet}
Jiangning Zhang, Chao Xu, Liang Liu, Mengmeng Wang, Xia Wu, Yong Liu, and Yunliang Jiang.
\newblock Dtvnet: Dynamic time-lapse video generation via single still image.
\newblock In \emph{Computer Vision--ECCV 2020: 16th European Conference, Glasgow, UK, August 23--28, 2020, Proceedings, Part V 16}, pages 300--315. Springer, 2020.

\end{thebibliography}
}
\clearpage
\setcounter{page}{1}
\maketitlesupplementary

\section{Expanded Approach Description}
We start this section by revisiting the mathematical formulation of conditional diffusion models, alongside an exploration of the self-attention mechanism and positional encoding inherent in transformers.

\subsection{Conditional Diffusion Models}

\textbf{Diffusion models} \cite{ho2020denoising} are generative models characterized by forward and reverse diffusion processes. The forward process destroys the data from the true distribution $\mathbf{x}_0 \sim q(\mathbf{x}_0)$ using a Markov chain to add Gaussian noise at each step: $q(\mathbf{x}_t|\mathbf{x}_{t-1}):= \mathcal{N}(\sqrt{1-\beta_t}\mathbf{x}_{t-1},\beta_t\mathbf{I})$, where $\beta_t$ is a small positive constant that represents the noise level, $t$ is the diffusion step, and $\mathbf{I}$ is the identity matrix.
Since the noise used at each step is Gaussian,  $q(\mathbf{x}_t|\mathbf{x}_0)$  can be obtained in  closed-form $q(\mathbf{x}_t|\mathbf{x}_0)=\mathcal{N}(\mathbf{x}_t;\sqrt{\alpha_t}\mathbf{x}_0,(1-\alpha_t)\mathbf{I})$, where $\alpha_t = \prod_{s=1}^t(1 - \beta_s)$.
The reverse process denoises $\mathbf{x}_t$ to recover $\mathbf{x}_0$, and is defined as: 
$ p_\theta(\mathbf{x}_{t-1}|\mathbf{x}_t):=\mathcal{N}(\mathbf{x}_{t-1};\mu_\theta(\mathbf{x}_t,t),\sigma_\theta(\mathbf{x}_t,t)\mathbf{I})$, where $\mu_\theta$ and $\sigma_\theta$ are approximated by a neural network. \cite{ho2020denoising} has shown that this reverse process can be trained by solving the optimization problem 
$\min_{\theta} \lVert \epsilon-\epsilon_{\theta}(\textbf{x}_t,t)\rVert ^2_2$, where $\epsilon_{\theta}$ is a trainable denoising function, and estimates the noise vector $\epsilon$ that was added to its noisy input $\mathbf{x}_t$. 
\newline \noindent \textbf{Conditonal Diffusion Models}
 Diffusion models are in principle capable of modelling conditional distributions of the form $p(\mathbf{x}|\mathbf{y})$, where $\mathbf{y}$ is a conditional input.  In the context of conditional diffusion models, the model learns to predict the noise added to the noisy input given a set of conditions, including the time step $t$ and the conditional inputs $\mathbf{y}$.  The reverse process in this case is defined as: $p_{\theta}(\mathbf{x}_{t-1}|\mathbf{x}_t,\mathbf{y}):=\mathcal{N}(\textbf{x}_{t-1};\mu_\theta(\textbf{x}_t,t | \mathbf{y}),\sigma_\theta(\textbf{x}_t,t | \mathbf{y})\textbf{I})$. 
The conditional diffusion model learns a network $\epsilon_{\theta}$ to predict the noise added to the noisy input $\mathbf{x}_t$ with 
\begin{equation}
     \mathcal{L}(\theta)=\mathbb{E}_{\textbf{x}_0\sim q(\textbf{x}_0),\epsilon \sim \mathcal{N}(0,I),t,\mathbf{y}} [\lVert \epsilon-\epsilon_\theta(\textbf{x}_t,t,y)\rVert ^2_2]
\end{equation}
where $ \textbf{x}_t=\sqrt{\alpha_t}\mathbf{x}_0 +(1-\alpha_t)\epsilon$, and $\mathcal{L}$ is the overall learning objective of the diffusion model.

\subsection{Transformers and positional encoding}

\noindent \textbf{Transformers} Recent advancements in Natural Language Processing (NLP) and Computer Vision (CV), exemplified by \textit{Transformers} \cite{vaswani2017attention,dosovitskiy2021an}, have garnered considerable success. In the realm of \textit{transformers}, each element in a sequence or every patch in an image is typically embedded into a vector or token. These tokens traverse through an architecture composed of a stack of Self-Attention (SA) and Multi-Layer Perceptron (MLP) modules, with the SA
mechanism standing out as their fundamental component.  The SA module is designed to capture long-range interactions among three types of inputs: queries \(\mathbf{Q}\), keys \(\mathbf{K}\), and values \(\mathbf{V}\), where the values are linearly combined according to the importance of each key representing each of the queries.

More concretely, let \(\mathbf{X} = [\mathbf{x}_1,\dots,\mathbf{x}_N] \in \mathbb{R}^{D\times N}\), denote a set of $N$ data tokens. Then, the queries, keys and values \(\mathbf{Q} = [\mathbf{q}_1,\dots,\mathbf{q}_{N}] \in \mathbb{R}^{d \times N}\),
\(\mathbf{K} = [\mathbf{k}_1,\dots,\mathbf{k}_{N}] \in \mathbb{R}^{d \times N}\), and
\(\mathbf{V} = [\mathbf{v}_1,\dots,\mathbf{v}_{N}] \in \mathbb{R}^{d \times N}\), are computed 
 through linear multiplication of the inputs  with learnable weights denoted by \(\mathbf{W}^{Q} \in \mathbb{R}^{D\times d}\), \(\mathbf{W}^{K}\in \mathbb{R}^{D\times d}\), and \(\mathbf{W}^{V}\in \mathbb{R}^{D\times d}\):
 \[
 \mathbf{Q} = \mathbf{W}^Q \mathbf{X} , \ 
 \mathbf{K} = \mathbf{W}^K \mathbf{X} , \ \text{and }
 \mathbf{V} = \mathbf{W}^V \mathbf{X}. 
 \]
 
 Subsequently, a similarity matrix, or the attention map, is calculated using the dot-product between  queries and keys. The normalized attention map,  is then employed for the weighted aggregation of the values:
\begin{equation}
    \text{{SA}}(\mathbf{X}) = \mathbf{V}   \text{{softmax}}\left(\frac{\mathbf{K}^T\mathbf{Q}}{\sqrt{d}}\right)
\end{equation}
where 
\begin{equation}
    \text{{softmax}}\left(\frac{\mathbf{K}^T\mathbf{Q}}{\sqrt{d}}\right)_{i,j}
=  \frac{\exp(\mathbf{k}_i^T \mathbf{q}_j / \sqrt{d})}{\sum_i \exp(\mathbf{k}_i^T \mathbf{q}_j / \sqrt{d})}
\end{equation}
An essential characteristic of SA is its lack of awareness of positional information in the input\cite{xiao2023random}. In other words, the output's content remains independent of the input order:
\begin{equation}
\mathcal{S}(\text{{SA}}(\mathbf{X})) = \text{{SA}}(\mathcal{S}(\mathbf{X}))
\end{equation}
where \(\mathcal{S}(\mathbf{X}) = [\mathbf{x}_{\pi_1},\dots,\mathbf{x}_{\pi_N}]\) denotes a shuffle operator on the tokens, with the left and right instances being the same permutation.
Positional information can be introduced into the tokens by incorporating  the use of positional encoding, which is added to the tokens before the SA and Multi-Layer Perceptron (MLP) layers.

As suggested in \cite{kim2021rethinking}, the information utilized by SA can be categorized into three types: 1) relative positional-based attention, 2) absolute position-based attention, and 3) contents-based attention. For solving temporal jigsaw puzzles, where positional information might be less informative, it is natural to prioritize the model's focus on content-based information.

\section{Implementation Details}

\begin{figure}[tb]
\centering
\includegraphics[width=0.5\textwidth]{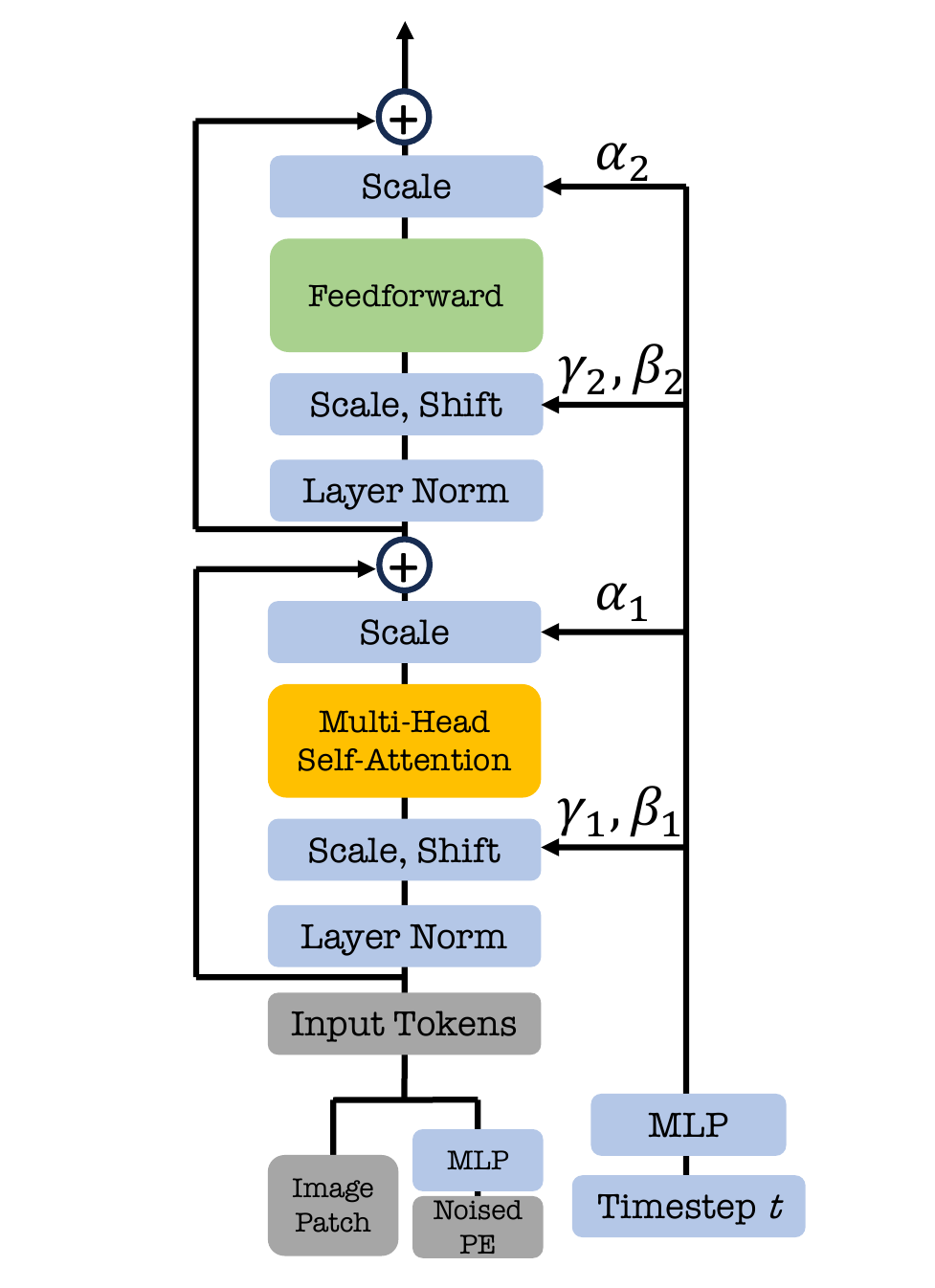}
\caption{Diffusion transformer block for image jigsaw puzzles.}\label{fig:dit}
\end{figure}

\begin{figure}[tb]
\centering
\includegraphics[width=0.5\textwidth]{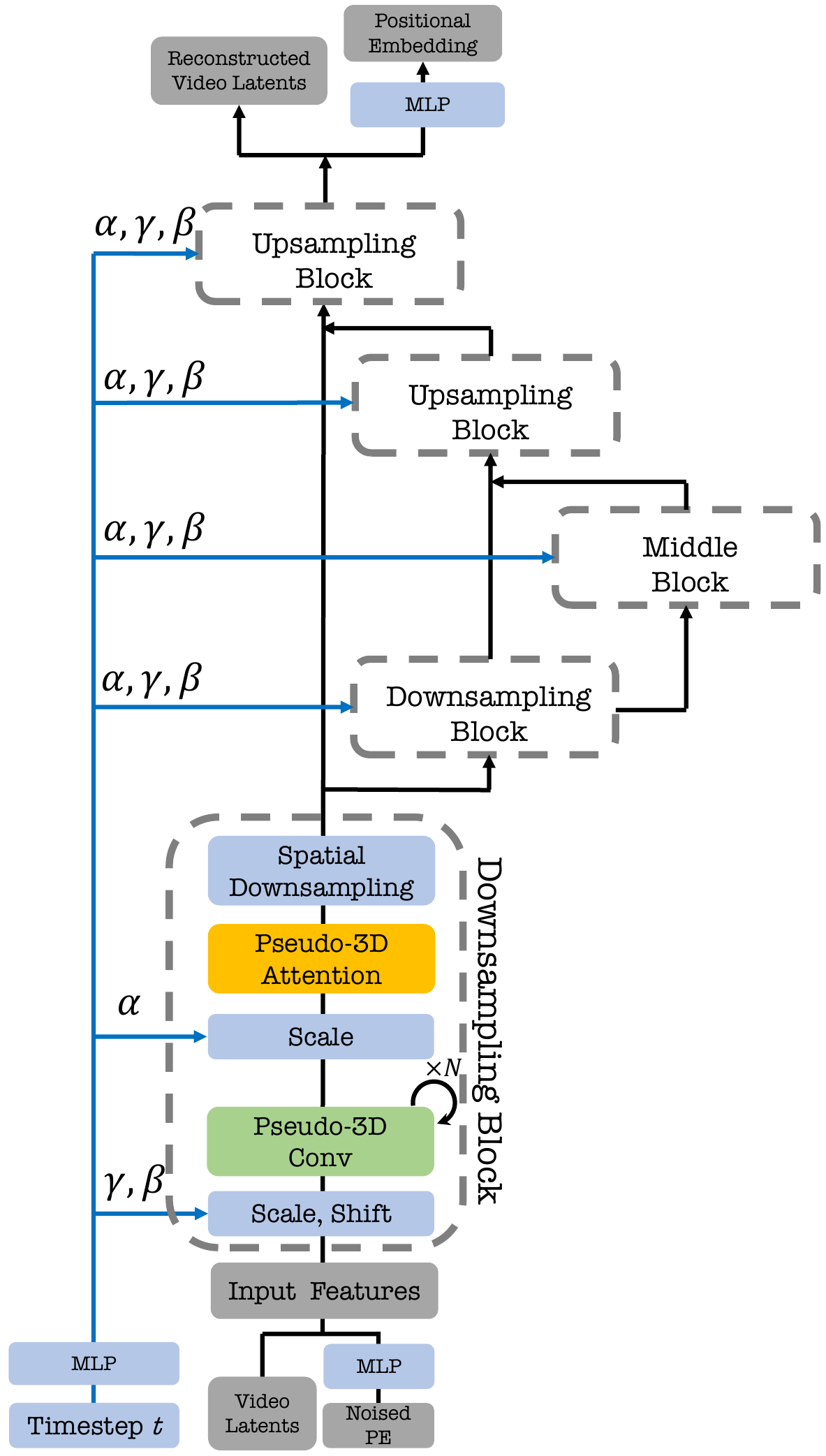}
\caption{Illustration of video model that we use for solving video jigsaw puzzles.}\label{fig:mav}
\end{figure}

\begin{figure*}[tb]
\centering
\includegraphics[width=1\textwidth]{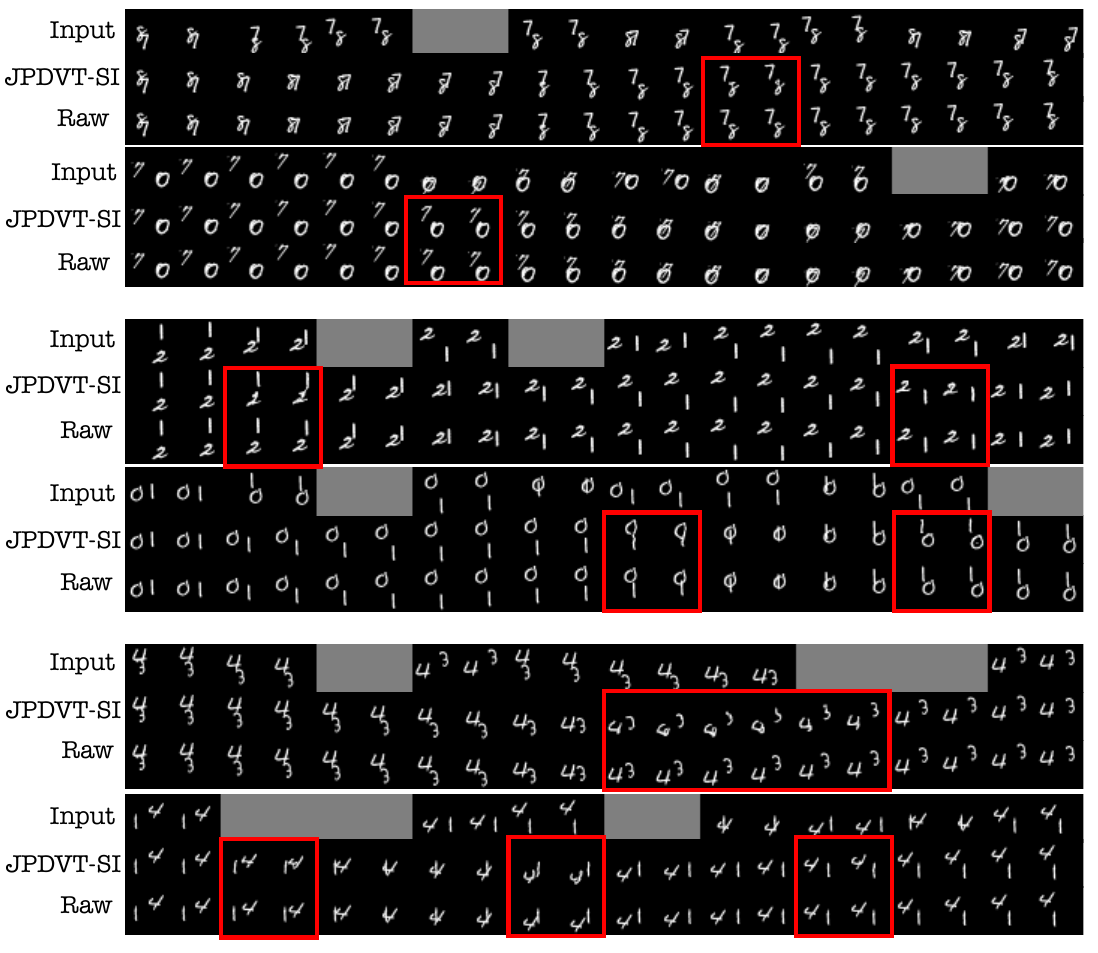}
\caption{The reshuffling and inpainting results on 10 pieces\textit{ MovingMNIST} experiment with different numbers of missing pieces. Frames highlighted within red boxes showcase the inpainting capabilities of our model.}\label{fig:mmnist_inpaint}
\end{figure*}

\begin{figure*}[tb]
\centering
\includegraphics[width=1\textwidth]{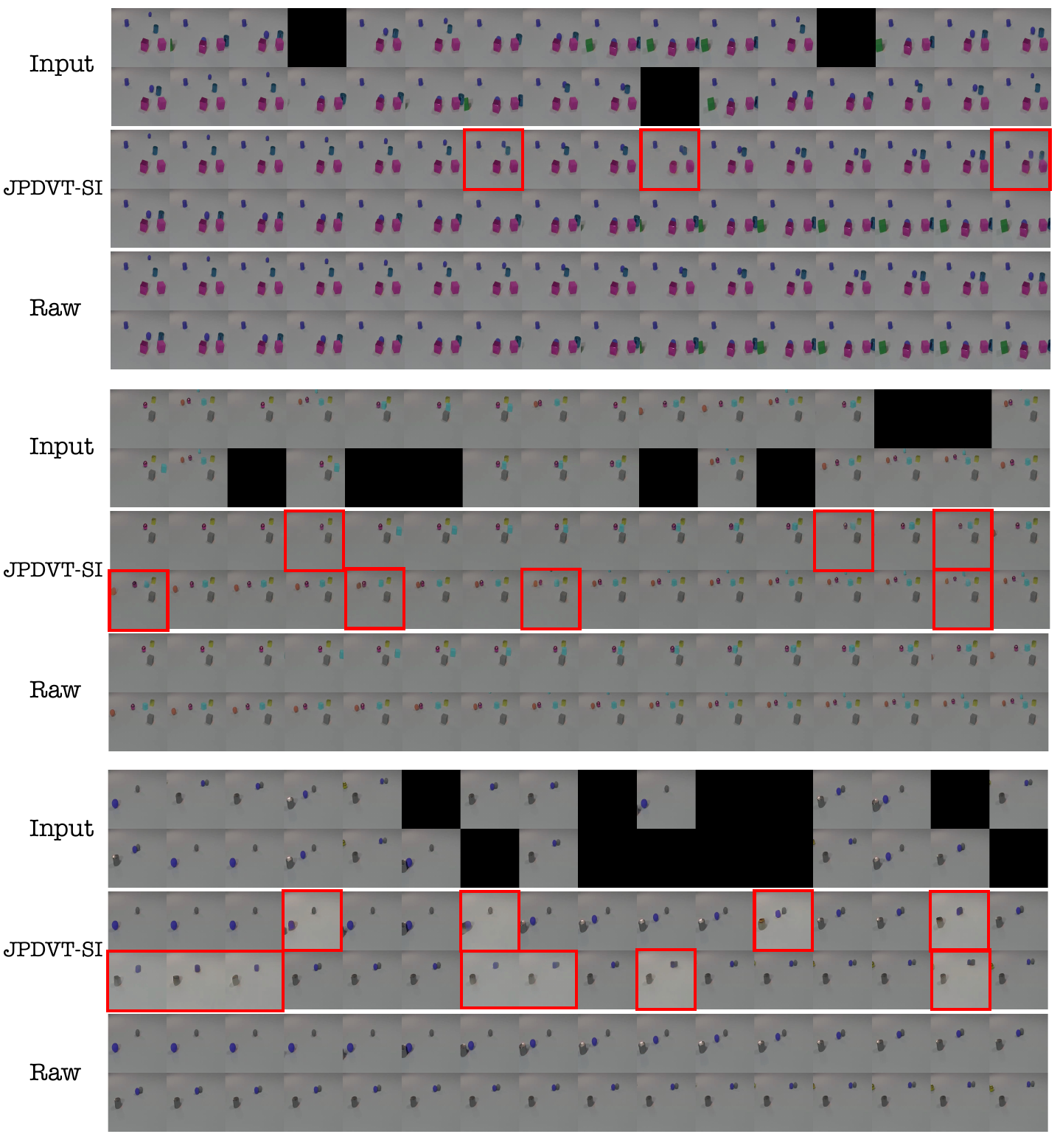}
\caption{The reshuffling and inpainting results on 32 pieces\textit{ CLEVRER} experiment with different numbers of missing pieces. Frames highlighted within red boxes showcase the inpainting capabilities of our model.}\label{fig:clvr_inpaint}
\end{figure*}

\begin{figure*}[tb]
\centering
\vspace{-0.15cm}
\includegraphics[width=\linewidth]{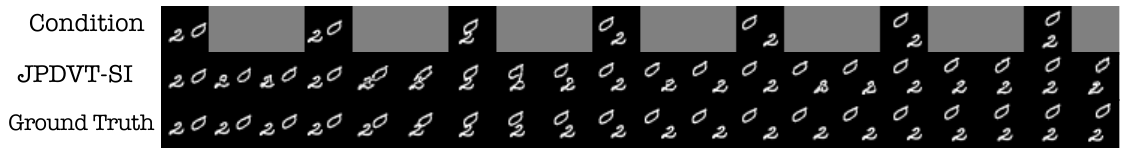}
\vspace{-0.5cm}
\caption{Video temporal super-resolution result.}\label{fig_video}
\vspace{-0.3cm}
\end{figure*} 

\subsection{Image Experiments}

\begin{table}
\centering
\caption{
Hyperparameters for our image diffusion transformer
}
\begin{tabular}{*2c}

Model & DiT     \\ 
\hline
Layers & 12 \\
Hidden dimension&  768  \\
MLP size & 3072 \\
Heads & 12\\
Patch size& 16 / 17 / 60 \\
\end{tabular} 

\vspace{-0.05in}
\label{tab:image_parameters}
\end{table}

\textbf{Diffusion Vision Transformer architecture.} We followed the standard ViT architecture, with extra Adaptive normalization layers added before and after MLP layers and multi-head attention layers, shown in Fig \ref{fig:dit}. Notably, our approach applies the diffusion model directly to image pixels rather than the latent space, providing a unique perspective on image understanding. In the context of image jigsaw puzzles without a gap, each image undergoes resizing and random cropping to a fixed size of $192 \times 192$ pixels. Within the DiT framework, the image is further segmented into $3 \times 3$ patches with $64 \times 64$ pixels resolution, subsequently embedded into 9 patch tokens. To explore the impact of eroded gaps in image jigsaw experiments, images are initially resized to $255 \times 255$ and then cropped into $3 \times 3$ patches with a resolution of $85 \times 85$ pixels. During training, a patch size of $64 \times 64$ is randomly selected, whereas in testing, a center crop method is employed to obtain image patches. The noised positional embedding undergoes processing through a Multilayer Perceptron (MLP) to align its dimension with that of the image tokens. Subsequently, the image tokens are added with the positional embedding tokens, forming an input that is fed into the diffusion transformers.

Our DiT models undergo training with a batch size of 256 over 300 epochs. A learning rate of $10^{-4}$ is employed, and the noised positional embedding undergoes diffusion through a linear schedule. The timestep $T$ is consistently set at 1000.

\noindent \textbf{ Predicted position encodings.} For a puzzle with $N$ pieces, we generate $N$ true position embeddings. Given a piece  diffusion generated positional embedding, we find the closest true embedding (using  $L_2$ distance) to assign the piece its final location. To avoid collisions, once a true position is used, it is removed as a candidate.   

\subsection{Video Experiments}

\textbf{Latent Diffusion Models.}
In our video models, we leverage the publicly available VAE encoder introduced in \cite{rombach2022high} to transition video frames from pixel space to latent space. Subsequently, we apply conditional diffusion models to the video latents. Each frame, initially sized at $256 \times 256 \times 3$, undergoes transformation into a latent feature map with dimensions $32 \times 32 \times 4$.
The positional embedding undergoes processing through an MLP layer before integration with the video latents. Our chosen architecture, as proposed by \cite{singer2022make}, is composed of four downsampling blocks, a middle block, and four upsampling blocks. Within each block, there are multiple pseudo-3D convolutional layers, a pseudo-3D attention layer, and a spatial downsampling/upsampling layer. An illustrative depiction of the architecture is presented in Figure \ref{fig:mav}, and detailed hyperparameters for each block are meticulously documented in Table \ref{tab:video_parameters}.

\begin{table}[t]
\centering
\caption{
Hyperparameters in the downsampling blocks for our video models
}
\begin{tabular}{*5c}

\#Block & $1^{st}$ & $2^{nd}$ & $3^{rd}$&$ 4^{th}$    \\ 
\hline
Hidden dimension& 64&128&256&512   \\
Feature map size &  32&16&8&4\\
\# P-3D conv layers &\multicolumn{4}{c}{3} \\
Attention head Dim & \multicolumn{4}{c}{64}\\
\# Attention heads & \multicolumn{4}{c}{8}\\
\end{tabular} 

\vspace{-0.05in}
\label{tab:video_parameters}
\end{table}

Our video models across various datasets were trained using a consistent batch size of 32. Each model underwent training for a total of 250,000 steps. We implemented a linear noise schedule, while keeping the total timestep, $T =1000$. The learning rate for the training process was set to $10^{-4}$. We trained our models on various datasets, each with a distinct temporal downsampling rate. Specifically, we applied downsampling factors of 1, 4, 2, and 2 for the MovingMNIST, CLEVRER, UCF101, and QST datasets, respectively. We used the same downsampling rate for the experiments run on the baselines.

In our experiments involving missing frames, we randomly sampled a varying number of missing elements, with the maximum set at 25\% of the total pieces. Gaussian noise was introduced to both the positional encoding matrix $\mathbf{L}$ and the matrix representing missing pieces $\mathbf{E}^m$. In our loss function, we employed a ratio of 0.8 to 0.2 to balance the loss term of missing frames and the loss term of the positional encoding.

\begin{figure}[htb]
\centering
\begin{adjustbox}{width=0.85\linewidth}
\includegraphics[width=\textwidth]{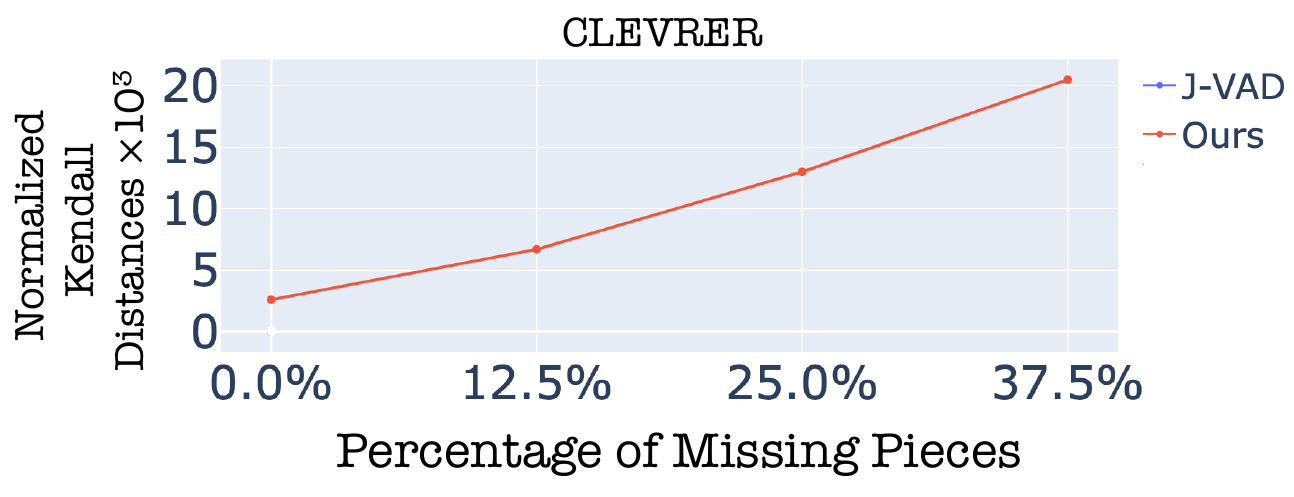}
 \end{adjustbox}
\caption{
CLEVRER dataset with 32 pieces, each with 1 frame. }\label{fig:video_plots_32p}
 \end{figure}
 
Additional results showcasing the outcomes of our video reshuffling and inpainting experiments are presented in Figures~\ref{fig:mmnist_inpaint} and \ref{fig:clvr_inpaint}. These figures vividly illustrate our models' proficiency in concurrently learning both visual context and positional information. Figure~\ref{fig:video_plots_32p} shows the results of Kendall distances of our proposed methods with different numbers of missing pieces with a total of 32 pieces on the CLEVRER dataset.

\noindent \textbf{Implementation of baselines.}
We conducted baseline experiments utilizing publicly available code from the following papers: \cite{dicle2016solving, xu2019self, wang2022video}. The specifics of the training configurations are outlined in Tables \ref{tab:vad_parameters} and \ref{tab:vcop_parameters}. In the case of experiments involving J-VAD \cite{wang2022video} and scenarios with more than 10 pieces, the models were trained for 2,000 epochs.

\begin{table}[t]
\caption{
Hyperparameters for experiments of J-VAD }
\begin{tabular}{cccc}
Dataset & MMNIST & CLEVRER & UCF101\\
\hline

optimizer & \multicolumn{3}{c}{AdamW} \\
learning rate &\multicolumn{3}{c}{1e-4} \\
momentum & \multicolumn{3}{c}{0.9}\\
weight decay & \multicolumn{3}{c}{5e-4}\\
image size &32 & 64 &112\\ 
batch size & 64 & 32 &8\\
epochs & 300 & 200 &200\\
\end{tabular}
\label{tab:vad_parameters}
\end{table}

\begin{table}[t]
\caption{
Hyperparameters for experiments on VCOP }

\begin{tabular}{*5c}
config & MMNIST & CLVR & UCF &QST\\
\hline
optimizer &\multicolumn{4}{c}{ SGD } \\
learning rate & \multicolumn{4}{c}{ 1e-3 } \\
momentum & \multicolumn{4}{c}{ 0.9} \\
weight decay & \multicolumn{4}{c}{ 5e-4 } \\
lr scheduler &\multicolumn{4}{c}{ RLRP }\\
batch size & 256 & 16&16&4 \\
image size &32 & 64 &112&128\\ 
\end{tabular}
\label{tab:vcop_parameters}
\vspace{-0.3cm}
\end{table}

\section{Downstream Task: Temporal Super-resolution}

The proposed framework can be directly applied to increase the temporal resolution of a given video, where  the missing puzzle pieces are the intermediate frames. Fig.~\ref{fig_video} shows an example where an input video with low sampling rate (top) was used to generate a video at a higher sampling rate (middle). 

\end{document}